\documentclass{article}

\PassOptionsToPackage{numbers,compress}{natbib}
\usepackage[preprint]{neurips_2026}

\usepackage[utf8]{inputenc}
\usepackage[T1]{fontenc}
\usepackage{microtype}
\usepackage{graphicx}
\usepackage{subcaption}
\usepackage{booktabs}
\usepackage{amsmath}
\usepackage{amssymb}
\usepackage{amsfonts}
\usepackage{amsthm}
\usepackage{mathtools}
\usepackage{nicefrac}
\usepackage{xcolor}
\usepackage{array}
\usepackage{multirow}
\usepackage{tabularx}
\usepackage{makecell}
\usepackage{enumitem}
\usepackage{xurl}
\usepackage{url}
\usepackage{placeins}
\usepackage{hyperref}
\usepackage[capitalize,noabbrev]{cleveref}

\theoremstyle{plain}

\theoremstyle{definition}

\theoremstyle{remark}

\DeclareRobustCommand{\TopK}{\ensuremath{\mathrm{Top}\text{-}\mathrm{K}}}
\DeclareRobustCommand{\TopKphi}{\ensuremath{\mathrm{Top}\text{-}\mathrm{K}{+}\phi}}

\DeclareRobustCommand{\NoSub}{\ensuremath{\mathrm{NoSub}}}
\DeclareRobustCommand{\FullSub}{\ensuremath{\mathrm{Sub}\text{-}\phi}}
\DeclareRobustCommand{\ExactSub}{\ensuremath{\mathrm{ExactSub}}}
\DeclareRobustCommand{\ExactNoSub}{\ensuremath{\mathrm{ExactNoSub}}}

\newcommand{\A}{\mathcal{A}}
\newcommand{\M}{\mathcal{M}}
\newcommand{\Kset}{\mathcal{K}}
\newcommand{\Rset}{\mathcal{R}}
\newcommand{\Eset}{\mathcal{E}}
\newcommand{\Gset}{\mathcal{G}}

\newcommand{\eps}{\varepsilon}
\newcommand{\Cmid}{C_{\mathrm{mid}}}
\newcommand{\rres}{\rho_{\mathrm{res}}}
\newcommand{\Loss}{\mathcal{L}}

\title{Residual-Mass Accounting for Partial-KV Decoding}

\author{%
  Yasuto Hoshi \qquad Daisuke Miyashita \qquad Jun Deguchi \\[0.5em]
  Kioxia Corporation \\
}

\begin{document}
\maketitle

\begin{abstract}
	We study a controlled partial-KV decoding setting in which exact unnormalized softmax contributions are computed for sink/tail anchors and a retrieved token set, while the remaining prefill tokens are represented by a residual estimate. We focus on the accounting rule after the query-dependent exact support has been selected, and use exhaustive \TopK{} only as an oracle selector, not as a deployable retrieval system. The proposed rule leaves the backbone language model and the exact-branch KV tensors unchanged. It builds fixed-size summary states $(S,u)$ from learned positive feature maps $\phi$, subtracts retrieved-token feature contributions to keep the exact and residual sets non-overlapping, and merges the estimated residual numerator and denominator with the exact branch under one normalization. At a 1\%  exact-support budget, our residual-completion method improves over the selection-only \TopK{} baseline on RULER and BABILong across frozen 1B and 3B Llama-3.2-Instruct backbones at all reported context lengths. In the 0.5--4\% exact-support budget sweeps, this trend largely persists. On LongBench, summarization results are mostly favorable, while multi-document QA is mixed. Attention-output diagnostics support retrieved-token subtraction as the partition-consistent accounting rule, while indicating that the main remaining error is imperfect learned-$\phi$ approximation of the unretrieved residual mass.
\end{abstract}

\section{Introduction}
\label{sec:introduction}

Decoder-only Transformers first run a \emph{prefill} pass to construct key--value (KV) states, then decode autoregressively using those cached states. At long context lengths, each decode step can reread a large prefix through the memory hierarchy, making decoding memory-traffic intensive \citep{gholami2024ai_memory_wall}. This pressure is also relevant when prefix KV states are reused across prompts or requests \citep{gim2024promptcache,yang2025learned,cheng2025lmcache}, or moved across GPU, CPU, storage, and network tiers \citep{sheng2023flexgen,cheng2025lmcache}.

We study a query-aware retrieval form of partial-KV decoding \citep{tang2024quest,lee2024infinigen,liu2024retrievalattention}, rather than KV-cache eviction \citep{zhang2023h2o,liu2023scissorhands}. For each decode query, the method exactly evaluates fixed sink/tail anchors and a query-dependent retrieved subset of the prefill KV cache; we call this set the \emph{exact support}. Instead of optimizing an end-to-end serving system, we isolate post-selection accounting: after the exact support is fixed, how should the remaining softmax mass---the unnormalized numerator and denominator contributions of tokens outside the support---be represented?

Selection-only sparse attention answers this accounting question by setting all contributions outside the exact support to zero and renormalizing over the support alone \citep{gupta2021topkattention,tang2024quest,liu2024retrievalattention}. This approximation is reliable when the omitted mass is negligible. Otherwise, tokens in the exact support receive inflated attention weights because omitted tokens are removed from both the unnormalized numerator and denominator before normalization.

Residual-mass accounting keeps the same exact support but changes the post-selection accounting rule. We implement residual-mass accounting with \TopKphi{}: during prefill, it builds fixed-size positive-feature summaries over the non-anchor mid-region. During decode, it computes exact unnormalized contributions for anchors and retrieved tokens. It then subtracts the retrieved-token feature contributions from the summaries to avoid double counting. Finally, it estimates the unnormalized numerator and denominator of the unretrieved residual, adds them to the corresponding exact terms, and applies a single normalization.

In experiments, we use exhaustive \TopK{} only as an oracle selector, not as a deployable retrieval system. This ensures that selection-only \TopK{} and \TopKphi{} use the same exact support---anchors plus retrieved tokens---and differ only in their post-selection accounting rule: \TopK{} discards the unretrieved residual tokens, whereas \TopKphi{} estimates their unnormalized numerator and denominator contributions before the final normalization.

This formulation separates two design problems. Selection asks which tokens enter the exact softmax computation. Accounting asks how the exact branch is combined with the remaining mass: whether omitted mass is dropped, retrieved tokens are double-counted, or the residual is added as unnormalized numerator and denominator terms and normalized once with the exact branch. This scope is complementary to concurrent sparse--linear architectures such as SPLA \citep{wang2026spla}: we do not convert the LM into a sparse-plus-linear attention model, but instead keep the exact support fixed and test how much omitted-mass accounting can recover.

\paragraph{Contributions.}
We make three contributions. First, we formulate fixed-support partial-KV decoding as residual-mass accounting and derive the subset-renormalization bias caused by dropping unretrieved mass and renormalizing over the selected support (\cref{sec:setup}). Second, we instantiate this rule in \TopKphi{}, a frozen-backbone residual-completion method that estimates the unretrieved softmax numerator and denominator with positive-feature summaries, subtracts retrieved-token contributions to keep the token partition non-overlapping, and merges exact and residual terms under one normalization (\cref{sec:method}). Third, using exhaustive \TopK{} as a control selector, we evaluate the rule on frozen 1B and 3B Llama-3.2-Instruct backbones and provide diagnostics that separate diffuse omitted-mass recovery, partition-consistent subtraction, and learned residual-calibration error (\cref{sec:natural_results,sec:stress_results,sec:accounting_analysis}).

\section{Related work}
\label{sec:related_work}

\paragraph{Partial-KV decoding and omitted mass.}
Prior work reduces decode-time KV reads by retaining or retrieving only part of the prefix KV cache. StreamingLLM keeps attention sinks \citep{xiao2023streamingllm}; H2O retains heavy hitters \citep{zhang2023h2o}; SnapKV and PyramidKV allocate token- or layer-level budgets \citep{li2024snapkv,cai2024pyramidkv}; and Quest, InfiniGen, and RetrievalAttention select or prefetch query-/layer-conditioned KV tokens, pages, or blocks \citep{tang2024quest,lee2024infinigen,liu2024retrievalattention}. These methods primarily address which tokens or blocks should be read exactly. Our work fixes that support and studies how the unread softmax mass should be accounted for. Selection-only \TopK{} attention is effective when attention is concentrated, but it discards numerator and denominator mass when omitted attention is diffuse. MagicPIG also targets insufficiently sparse regimes using sampling-based estimators \citep{chen2024magicpig}; in contrast, our residual branch uses fixed-size learned feature summaries.

\paragraph{Feature maps and learned linearizations.}
Kernelized and linear-attention methods approximate or replace softmax attention with additive feature-map summaries \citep{katharopoulos2020transformers,choromanski2021performer}. The exponential dot-product kernel admits classical and random-feature approximations \citep{smola2000dotproduct,kar2012random,wacker2024improved}, while learned methods such as Hedgehog, LoLCATs, LAWCAT, and related distillation approaches train linear or subquadratic attention variants to mimic softmax Transformers \citep{zhang2024hedgehog,zhang2024lolcats,liu-etal-2025-lawcat,goldstein2025radlads}. Our use of $\phi$ is narrower: the backbone LM, standard KV tensors, and exact softmax branch are left unchanged, and the learned maps are used only as an auxiliary estimator for unretrieved residual mass.

\paragraph{Hybrid sparse--compressive and sparse--linear residual attention.}
Hybrid memory designs combine explicit cache entries with compressed or recurrent representations. InfiniAttention adds compressive memory inside the Transformer block \citep{munkhdalai2024leave}. LoLA augments a linear-attention model with local and sparse global caches \citep{mcdermott2025lola}. The closest concurrent work is SPLA \citep{wang2026spla}, which combines exact block-sparse attention with residual linear attention and subtracts selected-block linear output from a global linear output. Our setting differs in the merge rule and architecture: we keep the original softmax LM and exact KV branch frozen, use the same selected tokens as \TopK{}, and estimate the unretrieved softmax numerator and denominator before one final normalization. Retrieved-token subtraction is therefore an accounting correction for a fixed token partition, not a block-selection rule or an architectural conversion to sparse-plus-linear attention.

\section{Setup and residual-mass bias}
\label{sec:setup}

During decoding, we approximate only prefill-segment attention.\footnote{Throughout this section, all quantities are for a single layer and a single query head; layer and head indices are omitted for readability.} Newly generated tokens remain exact. Let the prefill segment have length $L$, keys and values $(k_i,v_i)_{i=0}^{L-1}$, and query $q$. We use column vectors: $q,k_i\in\mathbb{R}^{d_h}$ and $v_i\in\mathbb{R}^{d_v}$. Here $d_h$ is the per-head query/key dimension and $d_v$ is the per-head value dimension; for the Llama backbones used in our experiments, $d_v=d_h$. Define the attention score $s_i$ and the full prefill-segment attention output $y_{\mathrm{full}}\in\mathbb{R}^{d_v}$ by

\begin{equation}
	s_i = \frac{q^\top k_i}{\sqrt{d_h}},
	\qquad
	y_{\mathrm{full}}
	=
	\frac{\sum_{i=0}^{L-1} \exp(s_i)v_i}{\sum_{i=0}^{L-1}\exp(s_i)}.
	\label{eq:full_attention}
\end{equation}

We partition the prefill indices into an exact anchor set $\A$ and a non-anchor \emph{mid-region} $\M$. The anchor set consists of initial sink tokens and recent tail tokens, while $\M$ is the portion between them. Formally,

\begin{equation}
	\A  = \{0,\ldots,n_{\mathrm{sink}}-1\}\cup\{L-n_{\mathrm{tail}},\ldots,L-1\}, \qquad
	\M  = \{n_{\mathrm{sink}},\ldots,L-n_{\mathrm{tail}}-1\}.
	\label{eq:partition}
\end{equation}

A selector returns a query-dependent retrieved set $\Kset(q)\subset\M$ of size $K$. We write $\Eset(q)$ for the exact support, and define its unnormalized numerator and denominator by

\begin{equation}
	\Eset(q)=\A\cup \Kset(q), \qquad
	\nu_{\Eset} = \sum_{i\in\Eset(q)} \exp(s_i)v_i, \qquad
	Z_{\Eset} = \sum_{i\in\Eset(q)} \exp(s_i).
	\label{eq:exact_terms}
\end{equation}

The selection-only \TopK{} attention output over the approximated prefill segment is
\begin{equation}
	y_{\TopK}=\frac{\nu_{\Eset}}{Z_{\Eset}}.
\end{equation}

Let $\Rset(q)=\M\setminus\Kset(q)$ denote the residual, or unretrieved, mid-region set, with
$\nu_{\Rset}=\sum_{i\in\Rset(q)}\exp(s_i)v_i$ and
$Z_{\Rset}=\sum_{i\in\Rset(q)}\exp(s_i)$.
Generated tokens are not included in $\Rset(q)$; at decode time, they remain exact and are added to the same unnormalized numerator and denominator as in \cref{eq:implemented_decode}.
If $Z_{\Rset}>0$ and $y_{\Rset}=\nu_{\Rset}/Z_{\Rset}$, then

\begin{equation}
	y_{\mathrm{full}}-y_{\TopK}
	=
	\frac{Z_{\Rset}}{Z_{\Eset}+Z_{\Rset}}\left(y_{\Rset}-y_{\TopK}\right).
	\label{eq:subset_bias}
\end{equation}

We refer to \cref{eq:subset_bias} as the \emph{subset-renormalization bias} of selection-only partial-KV attention. It is small when the omitted residual mass is small or when $y_{\Rset}\approx y_{\TopK}$, but can be large under diffuse attention, motivating residual completion.

\section{Method}
\label{sec:method}

\subsection{Positive feature maps and summary states}
\label{sec:phi_summaries}

We use $\phi$ for learned positive feature maps; \TopKphi{} denotes exact \TopK{} retrieval plus a $\phi$-parameterized residual branch, with cached summaries denoted by $S$ and $u$.

In practice, the feature maps are head-specific:
\begin{equation}
	\phi_{\mathrm q}^{\ell,h_{\mathrm q}},\phi_{\mathrm k}^{\ell,h_{\mathrm{kv}}}:\mathbb{R}^{d_h}\to\mathbb{R}_{>0}^{d_\phi},
	\qquad
	z(q,k)=\langle\phi_{\mathrm q}^{\ell,h_{\mathrm q}}(q),\phi_{\mathrm k}^{\ell,h_{\mathrm{kv}}}(k)\rangle.
\end{equation}
For readability, we omit layer and head superscripts. For grouped-query attention, $h_{\mathrm{kv}}$ denotes the KV head associated with query head $h_{\mathrm q}$ by the backbone's fixed head grouping.
If the feature maps exactly factorized the softmax kernel,
\begin{equation}
	\exp(q^\top k/\sqrt{d_h})=\langle\phi_{\mathrm q}(q),\phi_{\mathrm k}(k)\rangle
	\quad\text{for all relevant }(q,k),
	\label{eq:ideal_phi}
\end{equation}
then the summary states would provide exact unnormalized softmax terms for any token subset represented through $\phi$.
With fixed exact support, mismatch comes from finite-dimensional learned-$\phi$ residual-estimation error. The backbone LM is frozen, $\phi$ is used only after exact retrieval to estimate the residual, and all main experiments use $d_\phi=64$. The feature-map architecture and training details are collected in Appendix~\ref{sec:app_phi_training}.

During prefill, we build fixed-size summary states from the mid-region KV states:
\begin{equation}
	S_{\M} = \sum_{i\in\M} v_i\phi_{\mathrm k}(k_i)^\top\in\mathbb{R}^{d_v\times d_\phi},
	\qquad
	u_{\M} = \sum_{i\in\M}\phi_{\mathrm k}(k_i)\in\mathbb{R}^{d_\phi}.
	\label{eq:natural_cache}
\end{equation}
These states are built once per prefix, layer, and KV head, then reused across decode steps. They estimate the mid-region unnormalized numerator and denominator as $\hat\nu_{\M}(q)=S_{\M}\phi_{\mathrm q}(q)\in\mathbb{R}^{d_v}$ and $\hat Z_{\M}(q)=\phi_{\mathrm q}(q)^\top u_{\M}$, respectively.

\subsection{Retrieved-token subtraction}
\label{sec:subtraction}

The summary states in \cref{eq:natural_cache} include all mid-region tokens, including the query-dependent retrieved set $\Kset(q)$, which is already handled exactly. Adding them unmodified to the exact branch would double-count retrieved mid-region tokens. We subtract their feature-space contribution:
\begin{equation}
	S_{\Rset}
	=
	S_{\M} - \sum_{i\in\Kset(q)} v_i\phi_{\mathrm k}(k_i)^\top,
	\qquad
	u_{\Rset}
	=
	u_{\M} - \sum_{i\in\Kset(q)}\phi_{\mathrm k}(k_i).
	\label{eq:subtraction}
\end{equation}
Because the retrieved tokens are already materialized for the exact branch, this correction does not require reading additional attention-side KV entries beyond the selected set. Our prototype recomputes their feature maps. A fused or cached implementation could avoid part of this overhead, but such optimization is outside the reported results. This subtraction is only a fixed-support accounting correction; it is not a block-selection rule or an architectural conversion of the LM.

\subsection{Single-normalization merge}
\label{sec:single_norm_merge}

The residual completion terms are $\hat\nu_{\Rset}(q)=S_{\Rset}\phi_{\mathrm q}(q)$ and $\hat Z_{\Rset}(q)=\phi_{\mathrm q}(q)^\top u_{\Rset}$. For the prefill segment analyzed in \cref{sec:setup}, the hybrid output is
\begin{equation}
	y_{\TopKphi}
	=
	\frac{\nu_{\Eset}+\hat\nu_{\Rset}}{Z_{\Eset}+\hat Z_{\Rset}}.
	\label{eq:hybrid_merge}
\end{equation}

Under the ideal factorization in \cref{eq:ideal_phi}, the subtraction in \cref{eq:subtraction} gives $\hat\nu_{\Rset}=\nu_{\Rset}$ and $\hat Z_{\Rset}=Z_{\Rset}$, so \cref{eq:hybrid_merge} recovers the full prefill-segment attention in \cref{eq:full_attention}. We call \cref{eq:hybrid_merge} the \emph{residual-mass accounting rule}. Exact prefill contributions and estimated unretrieved prefill mass stay on the same unnormalized scale and are normalized once. Unlike separately normalized residual-output mixing with an ad hoc weight, this merge uses the softmax-consistent weights determined by $Z_{\Eset}$ and $\hat Z_{\Rset}$.

At decode time, already generated tokens are kept exact and added to the same normalization. If $\Gset$ denotes the generated-token prefix at the current step, with $\nu_{\Gset}=\sum_{i\in\Gset}\exp(s_i)v_i$ and $Z_{\Gset}=\sum_{i\in\Gset}\exp(s_i)$, the implemented output is
\begin{equation}
	y_{\mathrm{impl}}
	=
	\frac{\nu_{\Eset}+\hat\nu_{\Rset}+\nu_{\Gset}}
	{Z_{\Eset}+\hat Z_{\Rset}+Z_{\Gset}}.
	\label{eq:implemented_decode}
\end{equation}
Both equations use the same single-normalization merge, with generated tokens added as another exact branch.

\subsection{Training $\phi$}
\label{sec:training}

The backbone LM is frozen, and only the head-specific asymmetric maps $(\phi_{\mathrm q},\phi_{\mathrm k})$ are trained offline from full-attention teacher traces. The maps are one-block ReZero MLPs, and the objective combines temperature-scaled KL distillation with auxiliary top-band, false-positive, and residual log-partition penalties. Teacher traces are collected from long-form corpora including FineWeb, arXiv long-document summarization documents, and BIGPATENT \citep{penedo2024fineweb,cohan-etal-2018-discourse,sharma-etal-2019-bigpatent}. Full architecture, corpus, loss, optimization, and offline-cost details are consolidated in Appendix~\ref{sec:app_phi_training}.

Unless stated otherwise, every main evaluation uses one 64k-trained $\phi$ set per backbone LM. These maps are applied without retraining or adaptation to all evaluated context lengths up to 64k, and are not extrapolated beyond the context length used to train $\phi$. The controlled 4k design ablation in Appendix~\ref{sec:app_design_ablation} and the cross-length mismatch diagnostic in Appendix~\ref{sec:app_length_generalization} are the only exceptions.

\section{Experiments}
\label{sec:experiments}

\subsection{Models, tasks, and protocol}

We evaluate Llama-3.2-1B-Instruct and Llama-3.2-3B-Instruct \citep{grattafiori2024llama3herdmodels} as frozen backbone LMs up to 64k context length.

We use two evaluation families. For natural long-context generation, we follow the LongBench task taxonomy. We evaluate two summarization tasks (GovReport and QMSum; ROUGE-L) and two multi-document QA tasks (MuSiQue and HotpotQA; F1) \citep{bai-etal-2024-longbench}. For controlled stress tests, we use RULER and BABILong \citep{hsieh2024ruler,kuratov2024BABILong}. The main controlled table reports 4k, 16k, and 64k. Appendix~\ref{sec:app_stress_full} gives the full 4k/8k/16k/64k sweeps.

We evaluate every example provided by the corresponding lm-evaluation-harness task definitions \citep{eval-harness}: RULER has 500 examples for each of 13 tasks, BABILong uses QA1--QA5 for 4,996 examples, and each LongBench task uses 200 examples. All methods see the same inputs after truncation to 64k tokens, matching the supported length of the reported $\phi$ maps.

We use $n_{\mathrm{sink}}=4$, $n_{\mathrm{tail}}=16$, exhaustive \TopK{} retrieval over the mid-region, and total exact-support budget fractions $r\in\{0.5\%,1\%,2\%,4\%\}$. Sink/tail anchors count toward the budget; therefore,
\begin{equation}
	K=\max\left\{0,\left\lceil rL\right\rceil-|\A|\right\}
\end{equation}
mid-region tokens are retrieved. The exhaustive selector is a control condition. Because it scores the full mid-region, exhaustive \TopK{} is used only as an oracle control selector rather than as a deployable retrieval implementation. At a 1\% total exact-support budget, this yields \(K=21\), \(144\), and \(636\) retrieved mid-region tokens for 4k, 16k, and 64k contexts, respectively. ``Full exact'' is the unmodified backbone LM, ``\TopK{}'' is selection-only partial-KV attention, ``\TopKphi{}'' denotes our subtractive residual-completion method, and ``\TopKphi{} (\NoSub{})'' is an ablation that omits the retrieved-token subtraction in \cref{eq:subtraction}.
Accordingly, the budget fraction $r$ denotes the exact support included in the final attention merge, not an end-to-end memory-traffic budget.

\subsection{Natural long-context generation}
\label{sec:natural_results}

\Cref{tab:natural_main} reports the LongBench exact-support budget sweep for these two task families.

\begin{table*}[t]
	\centering
	\caption{LongBench exact-support budget-fraction sweep: summarization (GovReport and QMSum; ROUGE-L) and multi-document QA (MuSiQue and HotpotQA; F1). Sink/tail anchors count toward each percentage budget. All \TopKphi{} rows use the 64k-trained $\phi$. Bold marks the best partial-KV method within each task/backbone/budget block, including ties. ``Full exact'' is the unmodified backbone LM. ``\TopKphi{} (\NoSub{})'' intentionally omits retrieved-token subtraction.}
	\label{tab:natural_main}
	\setlength{\tabcolsep}{4.4pt}
	\renewcommand{\arraystretch}{1.1}
	\begin{tabular}{llcccccccc}
		\toprule
		     &                       & \multicolumn{4}{c}{\textbf{Llama-3.2-1B-Instruct}} & \multicolumn{4}{c}{\textbf{Llama-3.2-3B-Instruct}}                                                                                                       \\
		\cmidrule(lr){3-6}\cmidrule(lr){7-10}
		Task & Method                & 0.5\%                                              & 1\%                                                & 2\%            & 4\%            & 0.5\%          & 1\%            & 2\%            & 4\%            \\
		\midrule

		\multirow{4}{*}{GovReport}
		     & Full exact            & \multicolumn{4}{c}{0.280}                          & \multicolumn{4}{c}{0.331}                                                                                                                                \\
		     & \TopK                 & 0.266                                              & 0.271                                              & 0.275          & 0.276          & 0.323          & 0.324          & 0.320          & 0.321          \\
		     & \TopKphi{}            & 0.284                                              & 0.283                                              & \textbf{0.285} & \textbf{0.281} & 0.332          & \textbf{0.333} & 0.329          & \textbf{0.327} \\
		     & \TopKphi{} (\NoSub{}) & \textbf{0.288}                                     & \textbf{0.285}                                     & 0.282          & 0.278          & \textbf{0.336} & 0.330          & \textbf{0.331} & 0.325          \\
		\midrule

		\multirow{4}{*}{QMSum}
		     & Full exact            & \multicolumn{4}{c}{0.227}                          & \multicolumn{4}{c}{0.242}                                                                                                                                \\
		     & \TopK                 & 0.208                                              & 0.218                                              & 0.218          & 0.219          & 0.232          & 0.235          & 0.235          & 0.239          \\
		     & \TopKphi{}            & 0.216                                              & 0.215                                              & \textbf{0.224} & \textbf{0.227} & 0.234          & 0.234          & 0.235          & \textbf{0.241} \\
		     & \TopKphi{} (\NoSub{}) & \textbf{0.220}                                     & \textbf{0.220}                                     & 0.222          & 0.224          & \textbf{0.236} & \textbf{0.239} & \textbf{0.237} & \textbf{0.241} \\
		\midrule

		\multirow{4}{*}{MuSiQue}
		     & Full exact            & \multicolumn{4}{c}{0.178}                          & \multicolumn{4}{c}{0.112}                                                                                                                                \\
		     & \TopK                 & 0.186                                              & 0.192                                              & \textbf{0.189} & \textbf{0.191} & 0.096          & 0.104          & \textbf{0.112} & \textbf{0.119} \\
		     & \TopKphi{}            & 0.189                                              & 0.190                                              & \textbf{0.189} & 0.183          & 0.107          & \textbf{0.107} & 0.104          & 0.109          \\
		     & \TopKphi{} (\NoSub{}) & \textbf{0.190}                                     & \textbf{0.194}                                     & \textbf{0.189} & 0.184          & \textbf{0.110} & 0.106          & \textbf{0.112} & 0.115          \\
		\midrule

		\multirow{4}{*}{HotpotQA}
		     & Full exact            & \multicolumn{4}{c}{0.328}                          & \multicolumn{4}{c}{0.266}                                                                                                                                \\
		     & \TopK                 & 0.310                                              & 0.316                                              & 0.314          & 0.312          & \textbf{0.217} & \textbf{0.220} & 0.233          & 0.237          \\
		     & \TopKphi{}            & \textbf{0.315}                                     & \textbf{0.319}                                     & \textbf{0.316} & 0.310          & 0.207          & 0.219          & 0.226          & 0.237          \\
		     & \TopKphi{} (\NoSub{}) & 0.313                                              & 0.317                                              & 0.315          & \textbf{0.316} & 0.207          & 0.212          & \textbf{0.239} & \textbf{0.239} \\
		\bottomrule
	\end{tabular}
\end{table*}

LongBench results vary by task family, so this sweep is downstream evidence rather than mechanistic validation. Summarization is more favorable: subtractive \TopKphi{} improves over selection-only \TopK{} in 13 of 16 cells, ties in one, and has mean change $+0.006$. GovReport gains appear at every budget; QMSum is weaker but mostly non-negative relative to \TopK{}.

Multi-document QA is mixed. Across MuSiQue and HotpotQA, subtractive \TopKphi{} improves over \TopK{} in 6 of 16 cells, ties in two cells, and has mean change $-0.001$. MuSiQue improves for the 3B model at tighter budgets but loses at larger budgets. HotpotQA gives small gains for the 1B model at 0.5--2\%, but shows no consistent gain for the 3B model. Thus, we do not claim consistent improvement on LongBench multi-document QA.

Several partial-KV rows match or exceed Full exact; we treat these as downstream ROUGE/F1 artifacts from generation and string matching, not evidence of higher attention fidelity than full attention. Because \NoSub{} double-counts retrieved mid-region tokens, its occasional leading scores are best viewed as calibration compensation, not support for double counting. The controlled tests and attention-output diagnostics below more directly check missed mass and subtraction.

\subsection{Controlled long-context stress tests}
\label{sec:stress_results}

\Cref{tab:stress_1pct} reports 1\% total exact-support budget results for RULER and BABILong. These controlled tasks reduce free-form generation confounds, although final answer accuracy is still not a pure attention-fidelity metric. At this budget, \TopKphi{} improves over selection-only \TopK{} in every reported RULER and BABILong block. The full 0.5--4\% sweeps are in Appendix~\ref{sec:app_stress_full}. \NoSub{} can occasionally match or exceed \TopKphi{}. The diagnostics in \cref{sec:accounting_analysis} treat such cases as possible calibration compensation rather than evidence for a reliable double-counting rule.

\begin{table*}[t]
	\centering
	\caption{RULER and BABILong results at a 1\% total exact-support budget. RULER reports the official aggregate score. BABILong reports average accuracy over QA1--QA5. Bold marks the best partial-KV method within each benchmark/length/backbone block, including ties.}
	\label{tab:stress_1pct}
	\setlength{\tabcolsep}{4.4pt}
	\renewcommand{\arraystretch}{1.2}
	\begin{tabular}{llcccccc}
		\toprule
		          &                       & \multicolumn{3}{c}{\textbf{Llama-3.2-1B-Instruct}} & \multicolumn{3}{c}{\textbf{Llama-3.2-3B-Instruct}}                                                                     \\
		\cmidrule(lr){3-5}\cmidrule(lr){6-8}
		Benchmark & Method                & 4k                                                 & 16k                                                & 64k            & 4k             & 16k            & 64k            \\
		\midrule

		\multirow{4}{*}{RULER}
		          & Full exact            & 0.803                                              & 0.665                                              & 0.581          & 0.918          & 0.816          & 0.721          \\
		          & \TopK                 & 0.732                                              & 0.606                                              & 0.541          & 0.878          & 0.787          & 0.688          \\
		          & \TopKphi{}            & 0.753                                              & \textbf{0.617}                                     & \textbf{0.548} & \textbf{0.889} & \textbf{0.814} & \textbf{0.711} \\
		          & \TopKphi{} (\NoSub{}) & \textbf{0.754}                                     & \textbf{0.617}                                     & 0.543          & 0.886          & 0.809          & 0.708          \\
		\midrule

		\multirow{4}{*}{BABILong}
		          & Full exact            & 0.331                                              & 0.244                                              & 0.140          & 0.474          & 0.357          & 0.215          \\
		          & \TopK                 & 0.308                                              & 0.204                                              & 0.102          & 0.488          & 0.345          & 0.184          \\
		          & \TopKphi{}            & \textbf{0.324}                                     & \textbf{0.220}                                     & \textbf{0.122} & \textbf{0.497} & \textbf{0.353} & \textbf{0.197} \\
		          & \TopKphi{} (\NoSub{}) & 0.322                                              & \textbf{0.220}                                     & 0.115          & 0.493          & 0.349          & 0.191          \\
		\bottomrule
	\end{tabular}
\end{table*}

\section{Analysis: entropy and completion gains}
\label{sec:entropy_analysis}

By \cref{eq:subset_bias}, residual completion should help most when exact support misses non-negligible residual mass. We diagnose this with the entropy of teacher attention restricted to the non-anchor mid-region: high entropy means mass is diffuse over many tokens; consequently, small \TopK{} support is more likely to miss aggregate mass. Appendix~\ref{sec:app_entropy_figs} gives complementary mass-recovery curves and the estimated residual denominator fraction.

For the rows considered, the visible mid-region is the full prefill mid-region; therefore, $\M(q)=\M$. We define the mid-normalized teacher distribution and entropy as
\begin{equation}
	p_i =
	\frac{\exp(s_i)}
	{\sum_{j\in\M(q)}\exp(s_j)},
	\qquad
	H_{\mathrm{mid}}(q)
	=
	-\frac{1}{\log|\M(q)|}
	\sum_{i\in\M(q)} p_i \log p_i .
\end{equation}
Following prior use of attention entropy as a concentration diagnostic
\citep{clark2019bert,voita2019analyzing}, low $H_{\mathrm{mid}}$ indicates concentrated mid-region mass, while high $H_{\mathrm{mid}}$ indicates diffuse mass.

\begin{figure*}[t]
	\centering
	\includegraphics[width=0.7\linewidth]{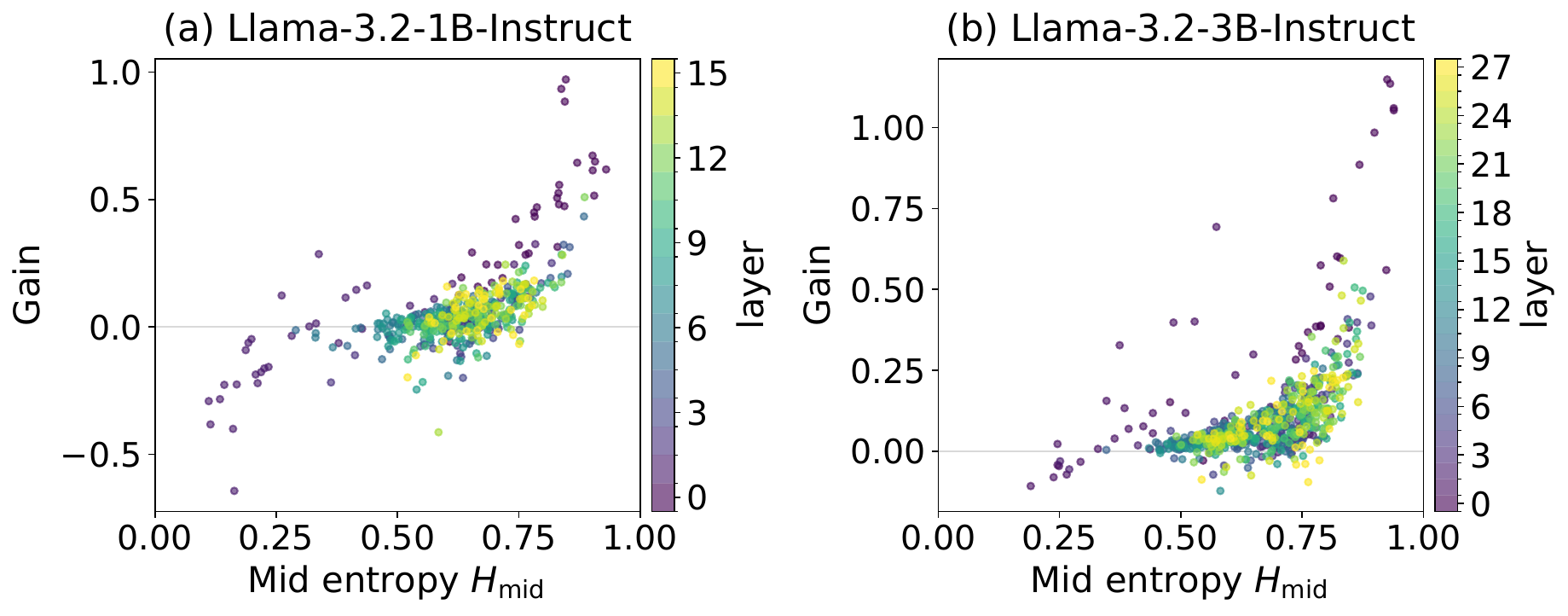}
	\caption{Head-wise completion gain versus mid-normalized attention entropy on Wikitext-64k. Each point corresponds to one layer--query-head. Gain is $e_{\TopK}-e_{\TopKphi}$, measured by relative $\ell_1$ attention-output error against full attention. Positive values indicate that subtractive residual completion reduces error relative to selection-only \TopK{}.}
	\label{fig:entropy_gain}
\end{figure*}

We measure relative $\ell_1$ attention-output error against full attention with $\eps=10^{-12}$:
\begin{equation}
	\mathrm{Rel}\text{-}\ell_1(\hat y,y_{\mathrm{full}})
	=
	\frac{\|\hat y-y_{\mathrm{full}}\|_1}{\|y_{\mathrm{full}}\|_1+\eps}.
\end{equation}

\Cref{fig:entropy_gain} compares head-wise completion gain with $H_{\mathrm{mid}}$ on Wikitext-64k. For Llama-3.2-1B-Instruct, the most pronounced negative gains appear in the low-entropy regime, especially around $H_{\mathrm{mid}}\lesssim 0.3$, although negative gains also occur at intermediate entropies. The learned residual branch then has little residual mass to recover and can introduce calibration error. For Llama-3.2-3B-Instruct, this low-entropy failure regime appears smaller. In both backbone LMs, the largest positive gains occur at high entropy and tend to increase with entropy.

These trends are consistent with the residual-mass decomposition in \cref{eq:subset_bias}. When mid-region teacher attention is diffuse, a small exact support leaves substantial aggregate mass outside the retrieved set, and \TopKphi{} can reduce subset-renormalization error by estimating unretrieved mass. When the distribution is sharp, \TopK{} often captures the dominant mid-region contribution, leaving less residual mass to recover and making learned residual-estimation error more visible. Appendix~\ref{sec:app_entropy_figs} provides teacher-side mass-recovery curves and same-$K$ breakdowns that support this explanation.

\section{Analysis: validating residual accounting}
\label{sec:accounting_analysis}

We check the accounting rule under a fixed retrieved set for the approximated prefill segment; generated tokens remain exact as in \cref{eq:implemented_decode}. For a query $q$,
\begin{equation}
	\Eset(q)=\A\cup\Kset(q),
	\qquad
	\Rset(q)=\M\setminus\Kset(q),
	\qquad
	\M=\Kset(q)\,\dot\cup\,\Rset(q).
	\label{eq:analysis_set_relation}
\end{equation}
Here $\Eset$ contains anchors and retrieved mid-region tokens, while $\M$ is the non-anchor mid-region split into retrieved and residual tokens. Since $\nu_{\Eset}=\nu_{\A}+\nu_{\Kset}$ and $\nu_{\M}=\nu_{\Kset}+\nu_{\Rset}$ (and likewise for $Z$), adding $\Eset$ and $\M$ double-counts $\Kset(q)$. The diagnostics therefore separate two issues: whether the accounting rule defines a valid non-overlapping partition, and how accurately the learned $\phi$ summaries estimate the residual terms under that partition.

\subsection{Oracle partition checks: subtraction and double counting}
\label{sec:oracle_partition_checks}

First consider oracle estimators using exact numerator and denominator terms. The exact-residual subtractive estimator is
\begin{equation}
	y_{\ExactSub}=\frac{\nu_{\Eset}+\nu_{\Rset}}{Z_{\Eset}+Z_{\Rset}}.
	\label{eq:exact_sub_oracle}
\end{equation}
Since $\Eset(q)=\A\cup\Kset(q)$ and $\M=\Kset(q)\dot\cup\Rset(q)$, \cref{eq:exact_sub_oracle} counts each contribution once and recovers full prefill-segment attention up to numerical precision (relative $\ell_1$ error approximately $10^{-7}$).

The exact-mid no-subtraction estimator is
\begin{equation}
	y_{\ExactNoSub}=\frac{\nu_{\Eset}+\nu_{\M}}{Z_{\Eset}+Z_{\M}}.
	\label{eq:exact_nosub_oracle}
\end{equation}
This is not a valid partition because $\Eset(q)$ already contains $\Kset(q)$ while $\M=\Kset(q)\dot\cup\Rset(q)$; consequently, retrieved mid-region tokens are counted twice. It is only an oracle double-counting diagnostic. In learned diagnostics, \FullSub{} denotes \TopKphi{} with subtraction, while \NoSub{} omits it.

\Cref{tab:exact_accounting} reports diagnostics on Wikitext and RULER Common Words Extraction (CWE) packs with Llama-3.2-1B-Instruct, using the 64k-trained $\phi$ maps. The 4k and 64k conditions use $K=21$ and $K=636$, respectively; both values are induced by the 1\% total exact-support budget after reserving the 20 sink/tail anchors.

\begin{table*}[t]
	\centering
	\caption{Residual-accounting diagnostics: mean relative $\ell_1$ attention-output error and residual log-partition calibration. \ExactNoSub{} uses the true whole-mid contribution without subtraction, double-counts $\Kset(q)$, and is diagnostic rather than a valid partition. The final column reports mean residual log-partition error $\log \hat Z_{\Rset}-\log Z_{\Rset}$; negative values indicate underestimation. Bold marks the lowest approximate method among \TopK{}, \NoSub{}, and \FullSub{}.}
	\label{tab:exact_accounting}
	\setlength{\tabcolsep}{4.4pt}
	\renewcommand{\arraystretch}{1.1}
	\begin{tabular}{lcrccccr}
		\toprule
		Corpus   & Length & $K$ & \TopK {\scriptsize$\downarrow$} & \NoSub {\scriptsize$\downarrow$} & \FullSub{} {\scriptsize$\downarrow$} & \ExactNoSub{} {\scriptsize$\downarrow$} & mean log-$Z$ err. \\
		\midrule
		Wikitext & 4k     & 21  & 0.267                           & 0.238                            & \textbf{0.191}                       & 0.156                                   & $-1.615$          \\
		CWE      & 4k     & 21  & 0.210                           & 0.290                            & \textbf{0.173}                       & 0.188                                   & $-2.378$          \\
		Wikitext & 64k    & 636 & 0.160                           & 0.147                            & \textbf{0.103}                       & 0.180                                   & $-1.843$          \\
		CWE      & 64k    & 636 & 0.184                           & 0.188                            & \textbf{0.151}                       & 0.183                                   & $0.485$           \\
		\bottomrule
	\end{tabular}
\end{table*}

Although \ExactNoSub{} has lower numerical error than \FullSub{} in some rows, it should not be treated as a valid baseline. It uses exact whole-mid terms that are unavailable to a partial-KV method, and because $\nu_{\Eset}$ already contains the retrieved tokens, adding $\nu_{\M}$ counts those tokens twice. Thus, \ExactNoSub{} is only a double-counting diagnostic, not an implementable method or an oracle for the desired non-overlapping partition. Among the approximate methods, \FullSub{} gives the lowest mean attention-output error in all four settings.

\subsection{Residual-estimation error in learned $\phi$ summaries}
\label{sec:learned_residual_calibration}

For a fixed partition, the remaining \FullSub{} error comes from replacing the true residual terms $(\nu_{\Rset},Z_{\Rset})$ with learned estimates $(\hat\nu_{\Rset},\hat Z_{\Rset})$. Under the ideal factorization in \cref{eq:ideal_phi}, subtraction would give $\hat\nu_{\Rset}=\nu_{\Rset}$ and $\hat Z_{\Rset}=Z_{\Rset}$; hence, \TopKphi{} would match full attention for the fixed support. The measured gap is therefore residual-estimation error, with $\log \hat Z_{\Rset}-\log Z_{\Rset}$ used as a denominator-scale diagnostic.

The final column of \Cref{tab:exact_accounting} reports mean residual log-partition error. It is negative in Wikitext-4k, Wikitext-64k, and CWE-4k, but mildly positive in CWE-64k. Thus, the residual denominator error does not have a consistent sign, so these diagnostics do not support correcting $\hat Z_{\Rset}$ with one global scaling factor. Subtraction removes the double-counting term by construction. After this valid partition is fixed, the remaining \FullSub{} error is residual-estimation error from the learned $\phi$ summaries, with the log-$Z$ diagnostics highlighting denominator calibration as an important target. Appendix~\ref{sec:app_residual_calibration} gives histogram-level views of the same calibration and \FullSub{}--\NoSub{} comparisons.

\section{Discussion and limitations}
\label{sec:discussion}

\paragraph{Scope and runtime.}
The contribution is an accuracy/accounting study of the fixed-support aggregation rule: exact attention on retrieved tokens, residual completion for unretrieved mid-region mass, retrieved-token subtraction, and a single normalization. Appendix~\ref{sec:app_runtime} reports prototype runtime. The unfused implementation is slower than Full exact SDPA and selection-only \TopK{}, and exhaustive \TopK{} is only a control selector to fix the exact support. For the same reason, SPLA \citep{wang2026spla} is complementary rather than a fixed-support baseline: it evaluates an adapted sparse-plus-linear architecture, whereas our experiments hold the selected tokens fixed and test only the accounting merge. The auxiliary $\phi$ modules add substantial parameter overhead: at $d_\phi=64$, 378.5M parameters for the 1B backbone and 559.3M for the 3B backbone; these weights are shared across requests and are separate from per-prefix summary states.

\paragraph{LongBench, metrics, and selector--accounting separation.}
Following the LongBench taxonomy, the natural-generation table separates ROUGE-L summarization from F1 multi-document QA. These are downstream generation metrics rather than direct attention-fidelity measures; therefore, the result should not be read as a uniform-improvement claim: summarization is mostly favorable, while multi-document QA is mixed. One plausible interpretation is that summarization benefits from recovering aggregate omitted context mass, whereas QA can depend more on whether the fixed exact support contains answer-bearing evidence and on small generation or string-match effects; the table alone does not identify which factor dominates. Because the main comparisons fix the exact support with exhaustive \TopK{}, selection-only \TopK{} and \TopKphi{} use the same selected tokens. Differences should therefore be read as accounting effects, not retrieval improvements. Future partial-KV systems should treat support retrieval and omitted-mass accounting as separate design problems. The controlled benchmarks and attention-output diagnostics test the merge rule more directly.

\paragraph{Calibration and generalization.}
Our learned finite-dimensional $\phi$ maps are approximate residual estimators, and residual denominator calibration mismatch is prominent in our diagnostics.
Appendix~\ref{sec:app_kernel_sanity} gives a kernel-sanity diagnostic reinforcing this intended role: learned $\phi$ scores estimate unretrieved residual mass rather than replacing exact retrieved pairs. Occasional task-level \NoSub{} wins can therefore be read as calibration compensation, not evidence that double counting is generally appropriate; the diagnostics in \cref{sec:learned_residual_calibration,sec:app_residual_calibration} indicate that this effect is dataset- and head-dependent. We evaluate only two Llama-3.2-Instruct backbone LMs. We also apply 64k-trained maps only up to their training length, avoid upward length extrapolation, and do not pair the method with an optimized retrieval system. Wider model families, longer extrapolation, better-calibrated residual objectives, and fused implementations remain future work.

\section{Conclusion}
We framed partial-KV decoding as residual-mass accounting: after anchors and retrieved tokens are evaluated exactly, the unretrieved tokens' unnormalized numerator and denominator contributions should be estimated and included in the same final normalization, rather than dropped as in selection-only \TopK{}, which normalizes over the exact support alone. A frozen-backbone implementation with learned positive feature maps, summary states, retrieved-token subtraction, and a single normalization improves over selection-only \TopK{} in the 1\% RULER/BABILong blocks. On LongBench, summarization results are favorable while multi-document QA is mixed. Diagnostics support subtraction as the partition-consistent accounting rule under the fixed-support decomposition studied here, and identify residual-mass calibration as a main remaining approximation challenge.

\clearpage
\bibliographystyle{plainnat}
\bibliography{kvcache_neurips26}

\clearpage

\appendix

\section{Appendix roadmap}
\label{sec:app_roadmap}

Appendix~\ref{sec:app_kernel_sanity} gives the kernel-sanity diagnostic and clarifies how the learned scores should be interpreted. Appendix~\ref{sec:app_phi_training} consolidates the feature-map model and all training details: background, ReZero architecture, teacher traces, corpus, loss, optimization, supported lengths, and offline cost. Appendix~\ref{sec:app_stress_full} gives the controlled stress-test exact-support budget sweeps for 4k/8k/16k/64k contexts. Appendix~\ref{sec:app_length_generalization} reports the cross-length $\phi$ mismatch diagnostic. Appendix~\ref{sec:app_residual_calibration} expands the residual-accounting calibration analysis, and Appendix~\ref{sec:app_design_ablation} reports the 4k feature-dimension and loss ablations. Appendix~\ref{sec:app_runtime} reports prototype runtime, summary-state size, and shared parameter overhead. Appendix~\ref{sec:app_entropy_figs} provides additional entropy, teacher-side mass-recovery, residual-denominator, and same-$K$ error analyses. Appendix~\ref{sec:app_reproducibility} lists implementation details and third-party assets.

\section{Kernel-sanity diagnostic}
\label{sec:app_kernel_sanity}

\begin{figure*}[t]
	\centering
	\includegraphics[width=\linewidth]{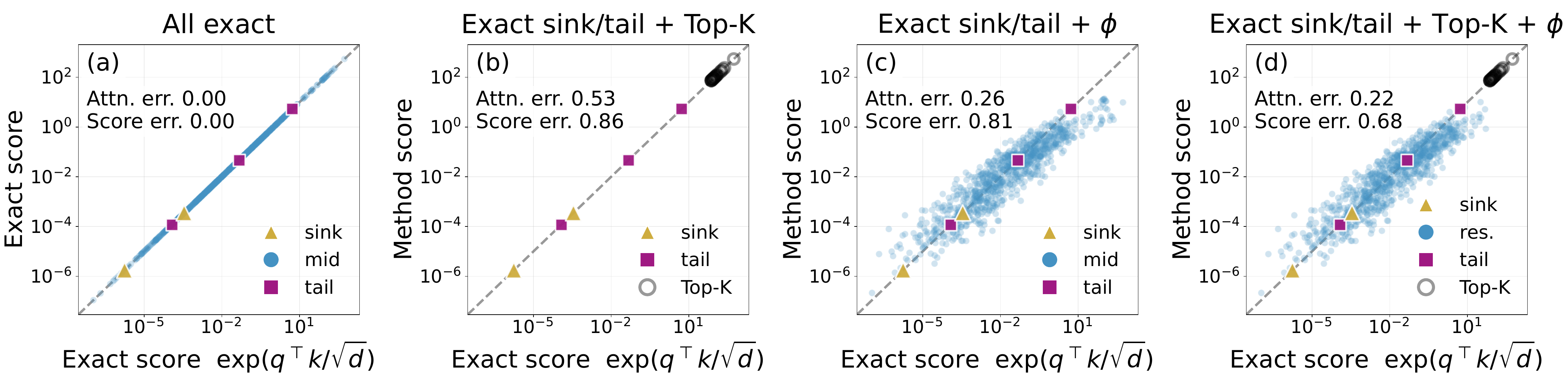}
	\caption{Exact softmax kernels and learned $\phi$ scores for a representative Llama-3.2-1B-Instruct head (layer~0, head~31) on Wikitext-64k using the 64k-trained maps. The diagnostic illustrates that learned $\phi$ scores are used only in the residual estimator.}
	\label{fig:kernel_sanity}
\end{figure*}

\paragraph{Role of the diagnostic.}
\Cref{fig:kernel_sanity} illustrates the intended separation between exact retrieved pairs and the learned residual branch. The learned $\phi$ scores are not meant to replace exact high-score pairs. Exact \TopK{} handles the retrieved support. The summary states built from $\phi$ estimate the unretrieved mass that selection-only normalization discards. Consistent with this role, adding the learned $\phi$ residual branch to exact sink/tail+\TopK{} reduces attention-output error in this diagnostic from 0.53 to 0.22.

\paragraph{What ``calibration error'' means here.}
In this paper, calibration error refers to mismatch in the unnormalized residual mass estimated from the summary states built with $\phi$. It does not refer to a separate probabilistic calibration procedure. Pairwise kernel errors in the learned scores can aggregate into numerator error $\hat\nu_{\Rset}-\nu_{\Rset}$ and denominator error $\hat Z_{\Rset}-Z_{\Rset}$ after subtraction. The scalar diagnostic used in the main text is the residual log-partition error $\log \hat Z_{\Rset}-\log Z_{\Rset}$. Negative values mean residual-mass underestimation, and positive values mean overestimation. The kernel-sanity plot (\Cref{fig:kernel_sanity}) should therefore be read as a check of the $\phi$-based residual estimator. It is not evidence that $\phi$ should replace exact retrieved pairs.

\FloatBarrier

\section{Feature-map model and training details}
\label{sec:app_phi_training}

This appendix describes the learned $\phi$ modules, including their architecture, training data, objective, optimization setup, supported lengths, and offline cost. The backbone LM is frozen throughout; $\phi$ is trained offline as an auxiliary residual estimator and is never used to update the backbone LM or to replace the exact retrieved-token branch.

\subsection{Feature-map background and modeling details}
\label{sec:app_feature_maps}

\paragraph{Symmetric feature maps as a reference point.}
The exponential dot-product kernel $\kappa(q,k)=\exp(q^\top k/\sqrt{d_h})$ has an exact, generally infinite-dimensional feature-map representation and admits randomized finite-dimensional approximations, including positive random features for softmax attention \citep{smola2000dotproduct,kar2012random,choromanski2021performer,wacker2024improved}. This provides the conceptual basis for additive summaries of keys and values.

\paragraph{Learned estimator.}
Our completion module uses learned positive kernel estimates
\begin{equation}
	z(q,k)=\langle \phi_{\mathrm q}(q),\phi_{\mathrm k}(k)\rangle>0,
\end{equation}
with feature maps $\phi_{\mathrm q},\phi_{\mathrm k}:\mathbb{R}^{d_h}\to\mathbb{R}_{>0}^{d_\phi}$. We allow $\phi_{\mathrm q}\neq\phi_{\mathrm k}$ for expressivity at fixed $d_\phi$. The method does not require the learned maps to be an exact feature-map factorization of the softmax kernel. It requires positivity and additive summary states to permit merging residual numerator and denominator estimates with the exact branch.

\paragraph{Why positivity matters.}
Because the residual branch contributes a denominator estimate, negative kernel estimates could make the normalizer invalid or unstable in this formulation. We therefore enforce positive feature outputs with exponentiated network outputs. Numerical clamps are used only to guard against finite-precision subtraction artifacts.

\subsection{Head-wise $\phi$ architecture}
\label{sec:app_phi_arch}

We learn feature maps per layer $\ell$, with $\phi_{\mathrm q}$ indexed by query head $h_{\mathrm q}$ and $\phi_{\mathrm k}$ indexed by KV head $h_{\mathrm{kv}}$:
\begin{equation}
	\phi_{\mathrm q}^{\ell,h_{\mathrm q}}:\mathbb{R}^{d_h}\to\mathbb{R}^{d_\phi},
	\qquad
	\phi_{\mathrm k}^{\ell,h_{\mathrm{kv}}}:\mathbb{R}^{d_h}\to\mathbb{R}^{d_\phi}.
\end{equation}

For a RoPE-applied input vector $x\in\mathbb{R}^{d_h}$ \citep{su2021roformer}, we use a one-block ReZero MLP. A stem maps $x$ to an internal width $d_{\mathrm{emb}}$, a gated residual MLP with GeLU nonlinearity updates the hidden vector, and a final linear layer followed by exponentiation produces positive features. All experiments use $d_{\mathrm{emb}}=512$:
\begin{align}
	g_0     & = W_s x+b_s,                               \\
	\Delta  & = W_2\operatorname{GeLU}(W_1 g_0+b_1)+b_2, \\
	g_1     & = g_0+\alpha\Delta,                        \\
	\phi(x) & =\exp(W_o g_1+b_o).
\end{align}
The architecture is shallow, but it is instantiated separately across layers and heads, so the full collection adds substantial parameter overhead. We disclose the total parameter counts in Appendix~\ref{sec:app_runtime}. The summary states built from $\phi_{\mathrm k}$ are stored per layer and KV head, while $\phi_{\mathrm q}$ is evaluated per query head at decode time.

\subsection{Training data, teacher traces, and support length}
\label{sec:app_training_pipeline}

For each sampled training text, we input a 64k-token sequence to the frozen teacher LM, run full attention, and collect the RoPE-applied query and key states together with teacher logits. From this 64k input, only 100 query positions are used for training; these positions are sampled uniformly at random from the final 33{,}000 tokens. For each selected query position $t$, the distillation target is the teacher attention distribution over all causally visible past keys for that query, i.e., all key positions $i\le t$, rather than only the ultimately retrieved \TopK{} keys. The logged tuples are therefore $(q,\{k_i\}_{i\le t},\{s_i\}_{i\le t})$ for selected query positions $t$. The student forms $z_i=\langle\phi_{\mathrm q}(q),\phi_{\mathrm k}(k_i)\rangle$ and $\hat s_i=\log z_i$. We do not replace attention inside the Transformer during $\phi$ training. Queries and keys therefore come from the full-attention teacher distribution, and the loss in Appendix~\ref{sec:app_loss} is applied only to the logged teacher and student logits.

Teacher traces are collected from long-form corpora, including FineWeb, arXiv long-document summarization documents, and BIGPATENT \citep{penedo2024fineweb,cohan-etal-2018-discourse,sharma-etal-2019-bigpatent}. We use the Hugging Face training sets of \texttt{HuggingFaceFW/fineweb}, \texttt{ccdv/arxiv-summarization}, and \texttt{NortheasternUniversity/big\_patent}. During $\phi$ training, optimization steps cycle across these three corpora, yielding a 1:1:1 corpus mixture over FineWeb, arXiv summarization, and BIGPATENT. For diagnostic evaluations on Wikitext, we use \texttt{wikitext-103-raw-v1}. Teacher attention is computed on the fly during $\phi$ training and consumed by the current optimization step; we do not store teacher-attention tensors to disk.

Unless stated otherwise, every main evaluation uses one set of head-specific $\phi$ maps trained offline at 64k context length for each backbone LM. The maps are applied unchanged to shorter contexts. We treat these maps as supported up to 64k and avoid upward extrapolation beyond their training length. The only exceptions are the controlled 4k design ablation in Appendix~\ref{sec:app_design_ablation} and the cross-length mismatch diagnostic in Appendix~\ref{sec:app_length_generalization}.

\subsection{Training objective}
\label{sec:app_loss}

For a fixed query, let $s_j$ be teacher logits and $\hat s_j=\log\langle\phi_{\mathrm q}(q),\phi_{\mathrm k}(k_j)\rangle$ be student logits. We shift by the teacher maximum $b=\max_j s_j$:
\begin{equation}
	r_j=s_j-b,
	\qquad
	\hat r_j=\hat s_j-b.
\end{equation}
With temperature $\tau$, define
\begin{equation}
	P=\operatorname{softmax}(r/\tau),
	\qquad
	\hat P=\operatorname{softmax}(\hat r/\tau).
\end{equation}
We set $\tau=10$ in the reported training runs and use
\begin{equation}
	\Loss_{\mathrm{KL}}=\tau^2\mathrm{KL}(P\|\hat P).
\end{equation}
We use Huber penalty $\rho_\delta$ and define a top band $\mathcal{B}=\{j:r_j\ge -\Delta\}$ and far region $\mathcal{F}=\{j:r_j< -\Delta\}$. The auxiliary terms are
\begin{align}
	\Loss_{\mathrm{top}} & = \mathbb{E}_{j\in\mathcal{B}}\rho_\delta(\hat r_j-r_j),                       \\
	\Loss_{\mathrm{fp}}  & = \mathbb{E}_{j\in\mathcal{F}}\rho_\delta\left(\max(\hat r_j+\Delta,0)\right), \\
	\Loss_Z              & = \rho_\delta\left(\max(\log\sum_j e^{\hat r_j}-\log\sum_j e^{r_j},0)\right).
\end{align}
The log-partition term is deliberately one-sided. It penalizes overestimation of the student partition function because, when the learned scores are used as a residual estimator, an overlarge residual denominator can give the residual branch too much weight in the single-normalization merge and interfere with the exact \TopK{} branch. We do not add a direct auxiliary penalty for partition-function underestimation: this is a conservative bias toward keeping the exact retrieved-token contributions authoritative. Underestimation is still indirectly constrained by the KL and top-band terms, but the explicit log-partition shaping is designed to prevent the residual estimate from getting in the way of the exact \TopK{} contribution.
We group these terms as
\begin{equation}
	\Loss_{\mathrm{aux}}
	=
	\lambda_{\mathrm{top}}\Loss_{\mathrm{top}}+\lambda_{\mathrm{fp}}\Loss_{\mathrm{fp}}+\lambda_Z \Loss_Z.
	\label{eq:app_aux_loss}
\end{equation}
Our full training objective is
\begin{equation}
	\Loss_{\mathrm{ours}}
	=
	\lambda_{\mathrm{KL}}\Loss_{\mathrm{KL}}+(1-\lambda_{\mathrm{KL}})\Loss_{\mathrm{aux}}.
	\label{eq:app_our_loss}
\end{equation}
Our default is $\tau=10$, $\lambda_{\mathrm{KL}}=0.99$, $\lambda_{\mathrm{top}}=1$, $\lambda_{\mathrm{fp}}=2$, $\lambda_Z=4$, $\Delta=12$, and $\delta=1$.

\subsection{Optimization and offline cost}
\label{sec:app_training_cost}

The $\phi$ modules are trained with AdamW, learning rate $10^{-3}$, weight decay $10^{-4}$, for 200k steps, using random seed 0. This cost is paid offline once per backbone LM rather than per request. After training, the same $\phi$ weights are reused for all prompts and exact-support budgets for that backbone LM.

\begin{table}[ht]
	\centering
	\caption{Training cost for the 64k-trained $\phi$ modules used in the experiments.}
	\label{tab:training_cost_64k}
	\renewcommand{\arraystretch}{1.1}
	\begin{tabular}{lcc}
		\toprule
		Backbone              & Steps / hardware            & Wall-clock \\
		\midrule
		Llama-3.2-1B-Instruct & 200k steps on 8$\times$H200 & 19 h       \\
		Llama-3.2-3B-Instruct & 200k steps on 8$\times$H200 & 35 h       \\
		\bottomrule
	\end{tabular}
\end{table}

\FloatBarrier

\section{Controlled stress-test sweeps}
\label{sec:app_stress_full}

\Cref{tab:ruler_sweep} reports the RULER sweep, and \cref{tab:babilong_sweep} reports the BABILong sweep. Baseline rows report the Full exact score for the corresponding context length. All \TopKphi{} rows use the 64k-trained $\phi$ maps except the separate design ablations in Appendix~\ref{sec:app_design_ablation}.

\begin{table*}[p]
	\centering
	\caption{RULER exact-support budget fraction sweep with one 64k-trained $\phi$ set per backbone LM. Sink/tail anchors count toward each percentage budget.}
	\label{tab:ruler_sweep}
	\setlength{\tabcolsep}{5pt}
	\renewcommand{\arraystretch}{1.1}
	\begin{tabular}{llcccccccc}
		\toprule
		       &                       & \multicolumn{4}{c}{\textbf{Llama-3.2-1B-Instruct}} & \multicolumn{4}{c}{\textbf{Llama-3.2-3B-Instruct}}                                                                                                       \\
		\cmidrule(lr){3-6}\cmidrule(lr){7-10}
		Length & Method                & 0.5\%                                              & 1\%                                                & 2\%            & 4\%            & 0.5\%          & 1\%            & 2\%            & 4\%            \\
		\midrule

		\multirow{4}{*}{4k}
		       & Full exact            & \multicolumn{4}{c}{0.803}                          & \multicolumn{4}{c}{0.918}                                                                                                                                \\
		       & \TopK                 & 0.547                                              & 0.732                                              & 0.762          & 0.784          & 0.698          & 0.878          & 0.898          & 0.912          \\
		       & \TopKphi{}            & 0.603                                              & 0.753                                              & \textbf{0.771} & \textbf{0.787} & \textbf{0.750} & \textbf{0.889} & \textbf{0.904} & \textbf{0.914} \\
		       & \TopKphi{} (\NoSub{}) & \textbf{0.605}                                     & \textbf{0.754}                                     & 0.769          & 0.784          & 0.745          & 0.886          & 0.902          & 0.913          \\
		\midrule

		\multirow{4}{*}{8k}
		       & Full exact            & \multicolumn{4}{c}{0.743}                          & \multicolumn{4}{c}{0.870}                                                                                                                                \\
		       & \TopK                 & 0.659                                              & 0.692                                              & 0.715          & 0.733          & 0.812          & 0.838          & 0.859          & 0.876          \\
		       & \TopKphi{}            & 0.687                                              & \textbf{0.705}                                     & 0.722          & 0.733          & \textbf{0.844} & \textbf{0.865} & \textbf{0.876} & \textbf{0.883} \\
		       & \TopKphi{} (\NoSub{}) & \textbf{0.688}                                     & \textbf{0.705}                                     & \textbf{0.723} & \textbf{0.734} & 0.841          & 0.863          & 0.874          & 0.880          \\
		\midrule

		\multirow{4}{*}{16k}
		       & Full exact            & \multicolumn{4}{c}{0.665}                          & \multicolumn{4}{c}{0.816}                                                                                                                                \\
		       & \TopK                 & 0.584                                              & 0.606                                              & 0.626          & 0.643          & 0.771          & 0.787          & 0.803          & 0.815          \\
		       & \TopKphi{}            & \textbf{0.602}                                     & \textbf{0.617}                                     & \textbf{0.631} & \textbf{0.646} & \textbf{0.800} & \textbf{0.814} & \textbf{0.823} & \textbf{0.824} \\
		       & \TopKphi{} (\NoSub{}) & \textbf{0.602}                                     & \textbf{0.617}                                     & \textbf{0.631} & 0.644          & 0.796          & 0.809          & 0.817          & 0.820          \\
		\midrule

		\multirow{4}{*}{64k}
		       & Full exact            & \multicolumn{4}{c}{0.581}                          & \multicolumn{4}{c}{0.721}                                                                                                                                \\
		       & \TopK                 & 0.529                                              & 0.541                                              & 0.553          & \textbf{0.563} & 0.676          & 0.688          & 0.701          & 0.711          \\
		       & \TopKphi{}            & \textbf{0.544}                                     & \textbf{0.548}                                     & \textbf{0.556} & 0.562          & \textbf{0.710} & \textbf{0.711} & \textbf{0.715} & \textbf{0.718} \\
		       & \TopKphi{} (\NoSub{}) & 0.540                                              & 0.543                                              & 0.548          & 0.554          & 0.706          & 0.708          & 0.712          & 0.715          \\
		\bottomrule
	\end{tabular}
\end{table*}

\begin{table*}[p]
	\centering
	\caption{BABILong exact-support budget fraction sweep with one 64k-trained $\phi$ set per backbone LM. Sink/tail anchors count toward each percentage budget.}
	\label{tab:babilong_sweep}
	\setlength{\tabcolsep}{5pt}
	\renewcommand{\arraystretch}{1.1}
	\begin{tabular}{llcccccccc}
		\toprule
		       &                       & \multicolumn{4}{c}{\textbf{Llama-3.2-1B-Instruct}} & \multicolumn{4}{c}{\textbf{Llama-3.2-3B-Instruct}}                                                                                                       \\
		\cmidrule(lr){3-6}\cmidrule(lr){7-10}
		Length & Method                & 0.5\%                                              & 1\%                                                & 2\%            & 4\%            & 0.5\%          & 1\%            & 2\%            & 4\%            \\
		\midrule

		\multirow{4}{*}{4k}
		       & Full exact            & \multicolumn{4}{c}{0.331}                          & \multicolumn{4}{c}{0.474}                                                                                                                                \\
		       & \TopK                 & 0.263                                              & 0.308                                              & 0.306          & 0.310          & 0.401          & 0.488          & \textbf{0.498} & 0.496          \\
		       & \TopKphi{}            & \textbf{0.298}                                     & \textbf{0.324}                                     & 0.321          & 0.319          & 0.431          & \textbf{0.497} & 0.494          & 0.495          \\
		       & \TopKphi{} (\NoSub{}) & 0.293                                              & 0.322                                              & \textbf{0.322} & \textbf{0.320} & \textbf{0.434} & 0.493          & 0.496          & \textbf{0.497} \\
		\midrule

		\multirow{4}{*}{8k}
		       & Full exact            & \multicolumn{4}{c}{0.289}                          & \multicolumn{4}{c}{0.406}                                                                                                                                \\
		       & \TopK                 & 0.255                                              & 0.254                                              & 0.257          & 0.265          & 0.409          & 0.418          & \textbf{0.421} & \textbf{0.420} \\
		       & \TopKphi{}            & \textbf{0.284}                                     & \textbf{0.273}                                     & 0.268          & \textbf{0.275} & 0.423          & \textbf{0.423} & \textbf{0.421} & \textbf{0.420} \\
		       & \TopKphi{} (\NoSub{}) & 0.283                                              & 0.272                                              & \textbf{0.270} & 0.274          & \textbf{0.426} & 0.420          & 0.416          & 0.415          \\
		\midrule

		\multirow{4}{*}{16k}
		       & Full exact            & \multicolumn{4}{c}{0.244}                          & \multicolumn{4}{c}{0.357}                                                                                                                                \\
		       & \TopK                 & 0.200                                              & 0.204                                              & 0.206          & 0.218          & 0.337          & 0.345          & 0.352          & \textbf{0.355} \\
		       & \TopKphi{}            & \textbf{0.229}                                     & \textbf{0.220}                                     & 0.218          & \textbf{0.223} & \textbf{0.358} & \textbf{0.353} & \textbf{0.354} & 0.353          \\
		       & \TopKphi{} (\NoSub{}) & 0.226                                              & \textbf{0.220}                                     & \textbf{0.219} & 0.222          & 0.352          & 0.349          & 0.352          & 0.354          \\
		\midrule

		\multirow{4}{*}{64k}
		       & Full exact            & \multicolumn{4}{c}{0.140}                          & \multicolumn{4}{c}{0.215}                                                                                                                                \\
		       & \TopK                 & 0.096                                              & 0.102                                              & 0.110          & 0.116          & 0.179          & 0.184          & 0.193          & 0.196          \\
		       & \TopKphi{}            & \textbf{0.123}                                     & \textbf{0.122}                                     & \textbf{0.121} & \textbf{0.127} & \textbf{0.196} & \textbf{0.197} & \textbf{0.200} & \textbf{0.203} \\
		       & \TopKphi{} (\NoSub{}) & 0.117                                              & 0.115                                              & 0.115          & 0.116          & 0.192          & 0.191          & 0.194          & 0.193          \\
		\bottomrule
	\end{tabular}
\end{table*}

In the RULER sweep, \TopKphi{} matches or improves over selection-only \TopK{} in 31 of 32 length--budget--backbone configurations. The largest gains typically appear at the tightest budgets, and margins generally shrink as the exact-support budget grows. The only exception is the 1B 64k/4\% setting, where \TopK{} is higher by 0.001. BABILong results are more variable, but at 64k both backbone LMs still benefit consistently from \TopKphi{}. The \NoSub{} variant is less stable. Overall, these sweeps support the main-text interpretation that residual completion is most beneficial in tight-budget regimes and remains beneficial at long context when missed mass is diffuse.

\FloatBarrier

\section{Cross-length \texorpdfstring{$\phi$}{phi} training mismatch}
\label{sec:app_length_generalization}

The main experiments use 64k-trained $\phi$ maps and evaluate them only at lengths up to 64k. To check whether shorter-trained maps can be safely extrapolated, we trained separate Llama-3.2-1B-Instruct $\phi$ maps at 4k, 8k, and 16k and evaluated them on RULER at a 1\% total exact-support budget. This is a separate mismatch diagnostic, not a replacement for the main RULER sweep. \Cref{tab:cross_length_mismatch} reports the diagnostic results.

\begin{table}[ht]
	\centering
	\caption{Cross-length mismatch diagnostic on RULER at 1\% total exact-support budget for Llama-3.2-1B-Instruct. Rows differ only in the maximum length used to train $\phi$; evaluation is performed at 4k, 8k, and 16k.}
	\label{tab:cross_length_mismatch}
	\renewcommand{\arraystretch}{1.1}
	\begin{tabular}{lccc}
		\toprule
		RULER @1\%               & 4k    & 8k    & 16k   \\
		\midrule
		Full exact               & 0.803 & 0.743 & 0.665 \\
		\TopK{}                  & 0.732 & 0.692 & 0.606 \\
		\TopKphi{} (4k-trained)  & 0.758 & 0.017 & 0.014 \\
		\TopKphi{} (8k-trained)  & 0.748 & 0.718 & 0.016 \\
		\TopKphi{} (16k-trained) & 0.759 & 0.704 & 0.634 \\
		\bottomrule
	\end{tabular}
\end{table}

The main failure mode is upward extrapolation beyond the maximum training length. The 4k-trained map collapses at 8k and 16k, and the 8k-trained map collapses at 16k. By contrast, the 16k-trained map remains usable at shorter or equal lengths. A practical default is therefore to avoid upward extrapolation. Accordingly, we treat a $\phi$ map trained at length $N$ as supported only up to $N$, and the main experiments use 64k-trained maps with 64k input truncation.

\section{Residual calibration and head-wise exceptions}
\label{sec:app_residual_calibration}

\begin{figure*}[p]
	\centering
	\includegraphics[width=\linewidth]{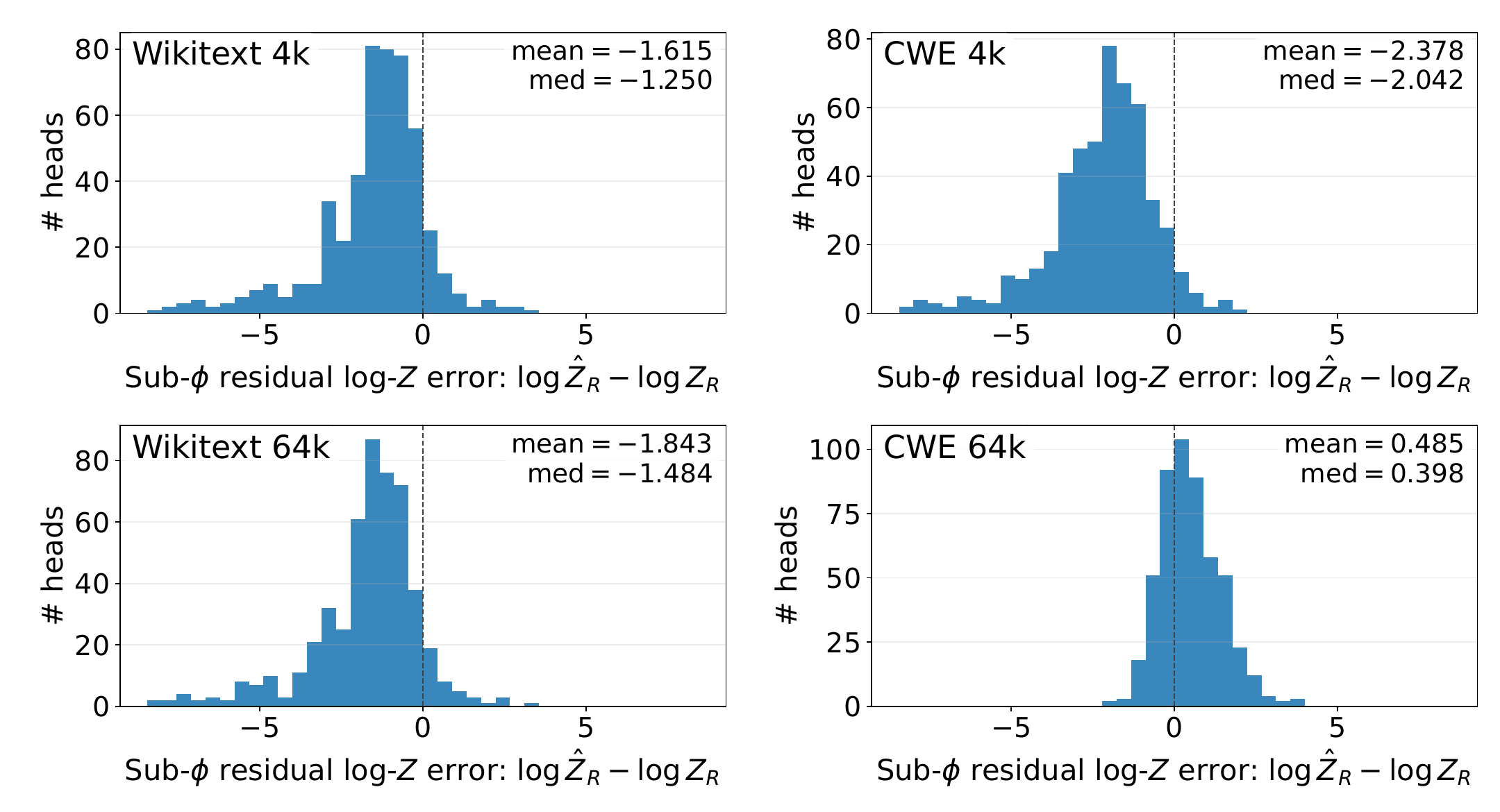}
	\caption{Residual denominator calibration for Llama-3.2-1B-Instruct. Histograms of $\log \hat Z_{\Rset}-\log Z_{\Rset}$ for \FullSub{} on Wikitext and CWE at 4k and 64k. Negative values indicate residual-mass underestimation.}
	\label{fig:logz_calibration}
\end{figure*}

\begin{figure*}[p]
	\centering
	\includegraphics[width=\linewidth]{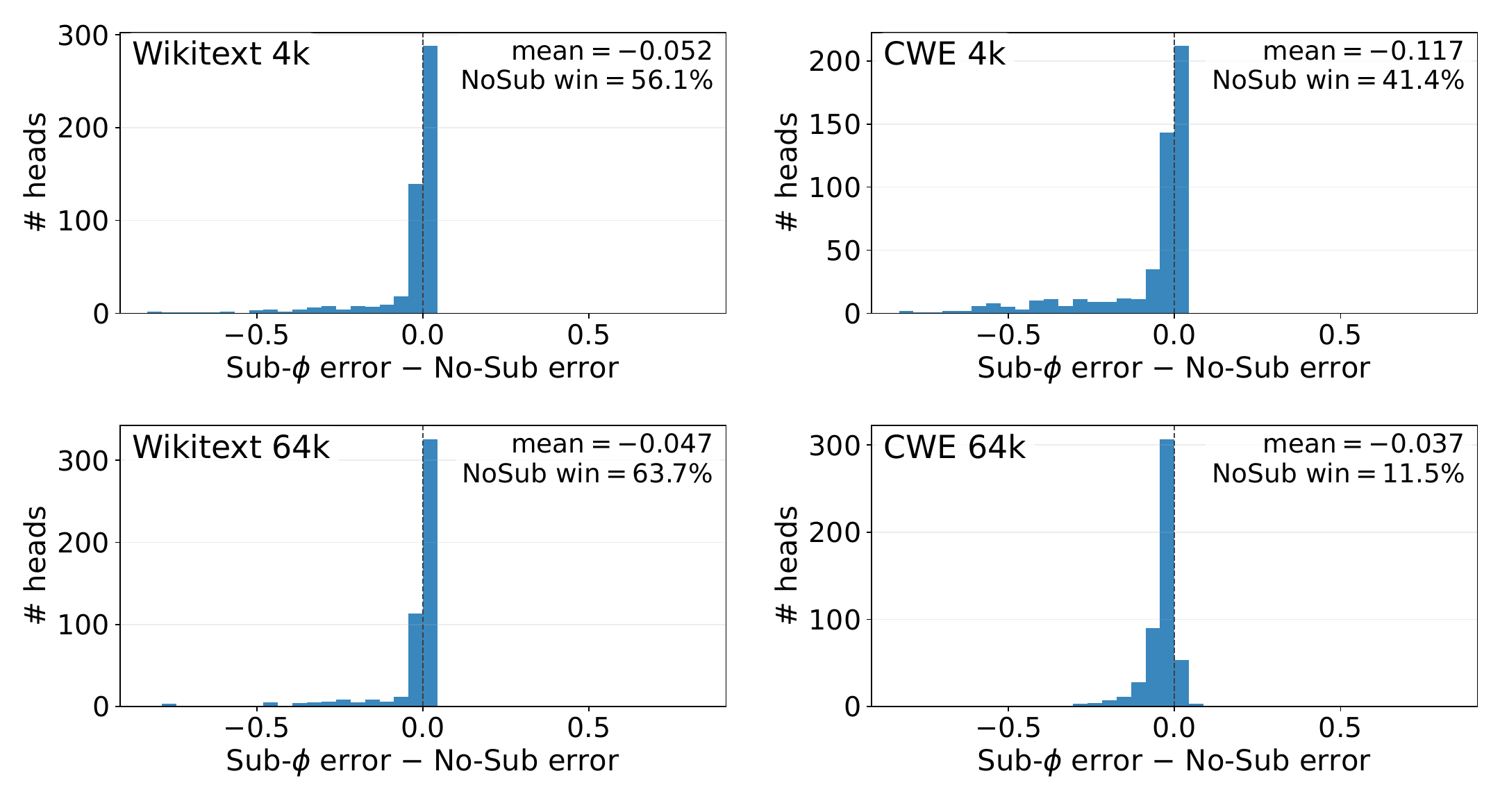}
	\caption{Head-wise \FullSub{}--\NoSub{} difference for Llama-3.2-1B-Instruct. Histograms of $\Delta=e_{\FullSub}-e_{\NoSub}$ across heads. Negative values favor \FullSub{}.}
	\label{fig:fullsub_nosub_delta}
\end{figure*}

This appendix expands the calibration analysis in \cref{sec:learned_residual_calibration} by visualizing the quantities summarized in \cref{tab:exact_accounting}. \Cref{fig:logz_calibration} shows the residual denominator calibration of the subtractive estimator, and \cref{fig:fullsub_nosub_delta} shows where head-wise exceptions arise when comparing \FullSub{} with \NoSub{}. These figures are distributional support for the main-text table rather than separate evaluation criteria.

As shown in \cref{fig:logz_calibration}, Wikitext-4k, Wikitext-64k, and CWE-4k have negative-centered residual log-partition errors, indicating that residual-mass underestimation is the dominant denominator-calibration pattern in these settings. CWE-64k is the exception: its distribution shifts slightly positive, indicating mild overestimation. Taken together, these distributions
show that calibration error does not have a universal sign; its sign and magnitude depend on corpus and length. This variability means that \NoSub{} does not provide a stable correction rule.

\Cref{fig:fullsub_nosub_delta} shows that the \FullSub{}--\NoSub{} differences are centered close to zero; therefore, subtraction does not dominate every head by a large margin. However, the mean difference is negative in all four settings, indicating that heads favoring \FullSub{} tend to favor it by larger margins. On Wikitext, \NoSub{} has lower error for slightly more than half of heads, but mostly by small amounts. On CWE, especially at 64k, \FullSub{} has lower error for a clear majority of heads. This pattern indicates unstable compensation: \NoSub{} can reduce error for some heads when residual calibration is mismatched, but it is not a partition-consistent decomposition of the attention mass.

\FloatBarrier

\section{4k design ablations: feature dimension and loss}
\label{sec:app_design_ablation}

\begin{figure*}[t]
	\centering
	\includegraphics[width=\linewidth]{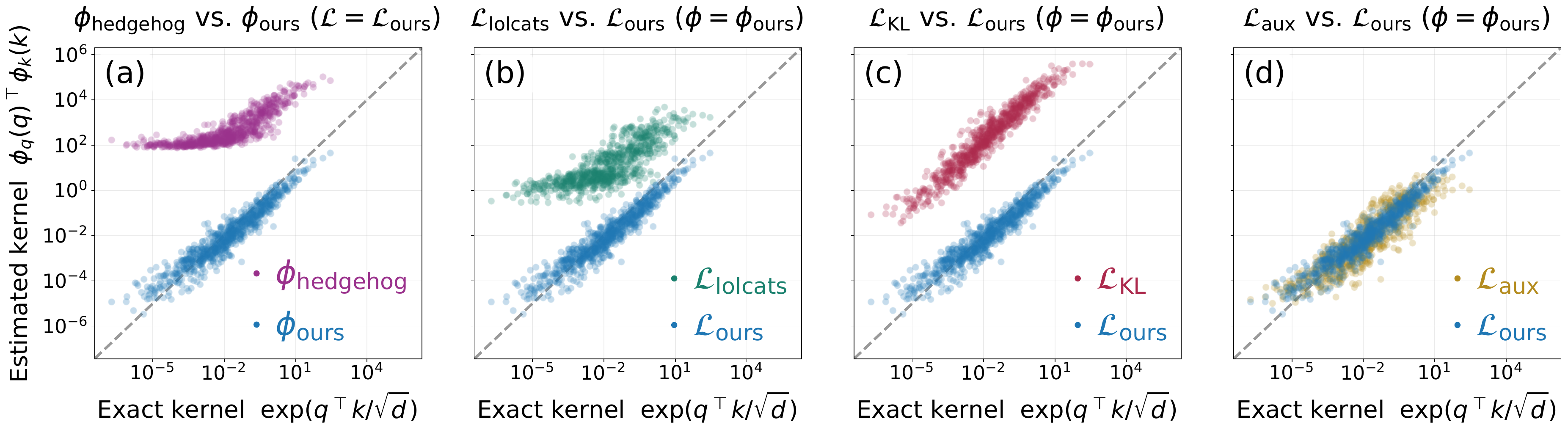}
	\caption{Mid-region kernel fidelity for \TopKphi{} completion on Wikitext-4k using Llama-3.2-1B-Instruct (layer~0, head~31). Panel (a) changes the feature map parameterization while keeping the loss fixed to $\Loss_{\mathrm{ours}}=\lambda_{\mathrm{KL}}\Loss_{\mathrm{KL}}+(1-\lambda_{\mathrm{KL}})\Loss_{\mathrm{aux}}$ (\cref{eq:app_our_loss}); panels (b)--(d) keep our feature map fixed and vary the loss ($\Loss_{\mathrm{lolcats}}$, KL only, or aux only). Points closer to the diagonal indicate more faithful approximation of unretrieved mid-region kernels.}
	\label{fig:app_ablation_kernel_fidelity}
\end{figure*}

This section reports controlled 4k design ablations using separately trained 4k $\phi$ maps, rather than the 64k-trained maps used in the main evaluations. \Cref{tab:dphi_ablation} varies the feature dimension, \cref{tab:loss_ablation} varies the loss/module choice, and \cref{fig:app_ablation_kernel_fidelity} visualizes representative kernel fidelity.
Panel (a) compares our feature map with the Hedgehog-style map at fixed loss, whereas panels (b)--(d) keep our feature map fixed and vary the loss.

``Hedgehog $\phi$ + our loss'' replaces our ReZero MLP feature map with a Hedgehog-style positive feature map \citep{zhang2024hedgehog}, while keeping the residual-completion training objective in Appendix~\ref{sec:app_loss}. For an input vector $x$, this diagnostic map computes
\begin{equation}
	z=xW+b,
	\qquad
	\phi(x)=\left[\exp(z),\exp(-z)\right].
	\label{eq:app_hedgehog_phi}
\end{equation}
``Our $\phi$ + LoLCATs loss'' keeps our feature map but replaces the objective with a LoLCATs-style attention-transfer loss \citep{zhang2024lolcats},
\begin{equation}
	\Loss_{\mathrm{lolcats}}
	=
	\operatorname{MSE}\left(
	\operatorname{out}_{\phi},
	\operatorname{out}_{\mathrm{exact}}
	\right),
	\label{eq:app_lolcats_attention_transfer_loss}
\end{equation}
where $\operatorname{out}_{\phi}$ and $\operatorname{out}_{\mathrm{exact}}$ are the feature-map and exact-attention outputs on the same logged teacher support. In the captions below, $\Loss_{\mathrm{ours}}=\lambda_{\mathrm{KL}}\Loss_{\mathrm{KL}}+(1-\lambda_{\mathrm{KL}})\Loss_{\mathrm{aux}}$ (\cref{eq:app_our_loss}) denotes the training objective used by the main $\phi$ maps. ``KL only'' and ``aux only'' retain only the corresponding terms.

The Hedgehog and LoLCATs rows test whether either replacing our feature map with the Hedgehog-style positive map or replacing our objective with the LoLCATs-style attention-transfer loss is sufficient for residual completion. Together with the KL-only and aux-only rows, these ablations are diagnostic swaps inside our residual-completion pipeline, not alternate accounting rules or evaluations of Hedgehog or LoLCATs in their native architectures or tuning regimes.

\begin{table}[ht]
	\centering
	\caption{Feature-dimension ablation with 4k-trained $\phi$ on Llama-3.2-1B-Instruct at a 1\% total exact-support budget. Sink/tail anchors count toward the 1\% budget.}
	\label{tab:dphi_ablation}
	\renewcommand{\arraystretch}{1.1}
	\begin{tabular}{lcc}
		\toprule
		Method       & RULER 4k       & BABILong 4k    \\
		\midrule
		Full exact   & 0.803          & 0.331          \\
		\TopK        & 0.732          & 0.308          \\
		$d_\phi=32$  & \textbf{0.758} & 0.335          \\
		$d_\phi=64$  & \textbf{0.758} & \textbf{0.339} \\
		$d_\phi=128$ & 0.753          & 0.334          \\
		\bottomrule
	\end{tabular}
\end{table}

\Cref{tab:dphi_ablation} shows that the 4k-trained maps are not strongly sensitive to increasing $d_\phi$ beyond 32 in this diagnostic. Both $d_\phi=32$ and $d_\phi=64$ reach 0.758 on RULER 4k, and $d_\phi=64$ gives the best BABILong 4k score, while $d_\phi=128$ is slightly worse on both benchmarks. Thus, 32--64 features appear sufficient in this 4k setting, and the improvement is not simply due to a larger feature state.

\begin{table}[ht]
	\centering
	\caption{Loss/module ablation with 4k-trained $\phi$ on Llama-3.2-1B-Instruct at a 1\% total exact-support budget. Sink/tail anchors count toward the 1\% budget. Hedgehog and LoLCATs refer to diagnostic swaps inside our residual-completion pipeline, not evaluations of those methods in their native settings.}
	\label{tab:loss_ablation}
	\renewcommand{\arraystretch}{1.1}
	\begin{tabular}{lcc}
		\toprule
		Method                     & RULER 4k       & BABILong 4k    \\
		\midrule
		Full exact                 & 0.803          & 0.331          \\
		\TopK                      & 0.732          & 0.308          \\
		\TopKphi{}                 & 0.758          & \textbf{0.339} \\
		Hedgehog $\phi$ + our loss & 0.016          & 0.079          \\
		Our $\phi$ + LoLCATs loss  & 0.018          & 0.079          \\
		Our $\phi$ + KL only       & 0.082          & 0.087          \\
		Our $\phi$ + aux only      & \textbf{0.762} & 0.329          \\
		\bottomrule
	\end{tabular}
\end{table}

\Cref{tab:loss_ablation} separates clear failure cases from competitive variants. The Hedgehog-style positive map and the LoLCATs-style attention-transfer loss both perform far below selection-only \TopK{} in this residual-completion setting, and KL-only also fails. By contrast, our full objective and the aux-only variant remain competitive: aux-only is slightly higher on RULER 4k, while the full objective is higher on BABILong 4k. Because aux-only retains the top-band, false-positive, and log-partition terms, these results suggest that auxiliary shaping is important for the single-normalization merge, while the exact KL--auxiliary balance is task-dependent in this 4k diagnostic.

\Cref{fig:app_ablation_kernel_fidelity} gives a kernel-level view of the same patterns in \cref{tab:dphi_ablation,tab:loss_ablation}. Panel (a) shows that replacing our map with the Hedgehog-style parameterization degrades mid-region kernel fidelity even under the same loss. Panels (b)--(d) show that, with the map fixed, LoLCATs loss and KL-only do not preserve the relevant mid-range scores well. Our full loss and aux-only remain closer to the diagonal. These ablations suggest that residual completion benefits from an appropriate model class and loss shaping in this setting: the Hedgehog-style map, the LoLCATs-style attention-transfer loss, and the KL-only objective were not sufficient, while aux-only retained much of the performance.

\section{Runtime, state, and parameter disclosure}
\label{sec:app_runtime}

These measurements disclose prototype overhead rather than an end-to-end speedup claim. In the unfused PyTorch implementation, the proposed method is slower than the Full exact SDPA baseline and the selection-only \TopK{} path. The measurements use an all-GPU prototype on 1$\times$H200, bf16, greedy decoding, batch size 1, and 32 generated tokens. Full exact uses PyTorch SDPA; \TopK{} and \TopKphi{} use the prototype path.
The reported $K$ values for 4k, 16k, and 64k are $21$, $144$, and $636$, respectively; these are the numbers of retrieved tokens from the mid-region induced by the 1\% total exact-support budget after reserving sink/tail anchors.
The summary states built from $\phi$ are constructed once during prefill for the input prefix. Decode-time overhead comes from query feature evaluation, retrieved-key feature recomputation, subtraction, and the final merge.

\Cref{tab:runtime_1b,tab:runtime_3b} report decode and end-to-end latency measurements for Llama-3.2-1B-Instruct and Llama-3.2-3B-Instruct. In this unfused prototype, \TopKphi{} is slower than \TopK{}. \Cref{tab:phi_param_count} summarizes the head-specific $\phi$ module parameter counts. These weights are loaded once per model instance and shared across requests. They are indexed by layer and head and are separate from the per-prefix summary state.

\begin{table}[ht]
	\centering
	\caption{Prototype latency disclosure for Llama-3.2-1B-Instruct. Each method column reports latency as per-step / end-to-end in milliseconds. The per-step value is averaged over generated tokens 2--32; end-to-end latency includes the full prefill pass plus 31 generation steps. $K$ is the number of retrieved mid-region tokens induced by the 1\% total exact-support budget after reserving sink/tail anchors.}
	\label{tab:runtime_1b}
	\renewcommand{\arraystretch}{1.1}
	\begin{tabular}{rrcccc}
		\toprule
		Context & $K$ & Full exact     & \TopK{}         & \TopKphi{}      & \TopKphi{} (\NoSub{}) \\
		\midrule
		4k      & 21  & 6.69 / 235.91  & 50.51 / 1594.37 & 76.85 / 2421.98 & 67.83 / 2142.50       \\
		16k     & 144 & 6.41 / 340.92  & 49.38 / 1673.12 & 84.41 / 2798.91 & 67.87 / 2288.69       \\
		64k     & 636 & 6.84 / 1430.63 & 52.04 / 2834.26 & 84.58 / 3996.32 & 66.12 / 3418.48       \\
		\bottomrule
	\end{tabular}
\end{table}

\begin{table}[ht]
	\centering
	\caption{Prototype latency disclosure for Llama-3.2-3B-Instruct, in the same format as \cref{tab:runtime_1b}.}
	\label{tab:runtime_3b}
	\renewcommand{\arraystretch}{1.1}
	\begin{tabular}{rrcccc}
		\toprule
		Context & $K$ & Full exact      & \TopK{}         & \TopKphi{}       & \TopKphi{} (\NoSub{}) \\
		\midrule
		4k      & 21  & 11.68 / 431.06  & 87.27 / 2774.16 & 141.85 / 4486.39 & 129.18 / 4093.23      \\
		16k     & 144 & 12.48 / 728.27  & 89.82 / 3125.23 & 148.23 / 5006.89 & 123.59 / 4242.39      \\
		64k     & 636 & 17.09 / 3380.97 & 88.54 / 5606.19 & 144.03 / 7571.53 & 122.12 / 6897.49      \\
		\bottomrule
	\end{tabular}
\end{table}

\begin{table}[ht]
	\centering
	\caption{Head-specific $\phi$ module parameter counts. ``Per head'' is the parameter count for one query-head or KV-head feature map. The total counts include all query-head maps and all KV-head maps.}
	\label{tab:phi_param_count}
	\setlength{\tabcolsep}{6pt}
	\renewcommand{\arraystretch}{1.1}
	\begin{tabular}{lccccrc}
		\toprule
		Backbone / $d_\phi$ & Per head  & q heads & KV heads & $\phi_{\mathrm q}$ total & $\phi_{\mathrm k}$ total & Total $\phi$ params \\
		\midrule
		1B / 32             & 575{,}010 & 512     & 128      & 294.4M                   & 73.6M                    & 368.0M              \\
		1B / 64             & 591{,}426 & 512     & 128      & 302.8M                   & 75.7M                    & 378.5M              \\
		1B / 128            & 624{,}258 & 512     & 128      & 319.6M                   & 79.9M                    & 399.5M              \\
		3B / 64             & 624{,}194 & 672     & 224      & 419.5M                   & 139.8M                   & 559.3M              \\
		\bottomrule
	\end{tabular}
\end{table}

For Llama-3.2-1B-Instruct with $d_\phi=64$ and head dimension 64, each per-prefix summary state consists of one $64\times64$ bf16 matrix $S$ plus one 64-dimensional bf16 vector $u$, per KV head/layer. This is approximately 66 KiB/layer and 1.03 MiB over 16 layers. The default 1B $d_\phi=64$ module has 378.5M shared parameters. The 3B $d_\phi=64$ module has 559.3M shared parameters.

\FloatBarrier

\section{Entropy, mass recovery, and residual-denominator diagnostics}
\label{sec:app_entropy_figs}

This appendix complements the entropy analysis in \cref{sec:entropy_analysis}. We keep two diagnostics separate because they answer different questions. The teacher-side mass-recovery curve asks how much of the true mid-region attention mass is captured by the exact \TopK{} set. The residual-denominator fraction asks how much mass the learned residual branch contributes after the final \TopKphi{} merge.

For a teacher attention row, define the mid-region denominator and the denominator captured by the top-$K$ mid-region tokens as
\begin{equation}
	Z_{\M}(q)=\sum_{j\in\M(q)}\exp(s_j),
	\qquad
	Z_{\Kset}(q;K)=\sum_{i\in \operatorname{TopK}_K(\M(q))}\exp(s_i).
\end{equation}
The teacher-side mid-region mass recovered by \TopK{} is
\begin{equation}
	\Cmid(K;q)
	=
	\frac{Z_{\Kset}(q;K)}{Z_{\M}(q)}
	=
	\frac{
		\sum_{i\in \operatorname{TopK}_K(\M(q))}\exp(s_i)
	}{
		\sum_{j\in\M(q)}\exp(s_j)
	} .
\end{equation}
Thus, $1-\Cmid(K;q)$ is the fraction of teacher mid-region mass missed by selection-only \TopK{}.

The learned residual branch is summarized by
\begin{equation}
	\rres(q)
	=
	\frac{\hat Z_{\Rset}(q)}{Z_{\Eset}(q)+\hat Z_{\Rset}(q)} .
\end{equation}
This is not the same quantity as $\Cmid$ or $1-\Cmid$: it uses the learned estimate $\hat Z_{\Rset}$, and its denominator includes the exact anchor and retrieved-token contributions. If $Z_{\A}(q)$ is the exact anchor mass and $Z_{\Rset}(q)=Z_{\M}(q)-Z_{\Kset}(q;K)$ is the true unretrieved mid-region mass, the corresponding teacher-side residual share is
\begin{equation}
	\rres^\star(q;K)
	=
	\frac{Z_{\Rset}(q)}{Z_{\A}(q)+Z_{\M}(q)}
	=
	\frac{(1-\Cmid(K;q))Z_{\M}(q)}{Z_{\A}(q)+Z_{\M}(q)} .
\end{equation}
Accordingly, $\Cmid$ measures coverage within the mid-region, while $\rres$ measures the implementation-side residual contribution after merging. Their connection is through the same missed residual mass, but $\rres$ can additionally reflect learned $\phi$ calibration error.

\begin{figure*}[ht]
	\centering
	\includegraphics[width=0.9\linewidth]{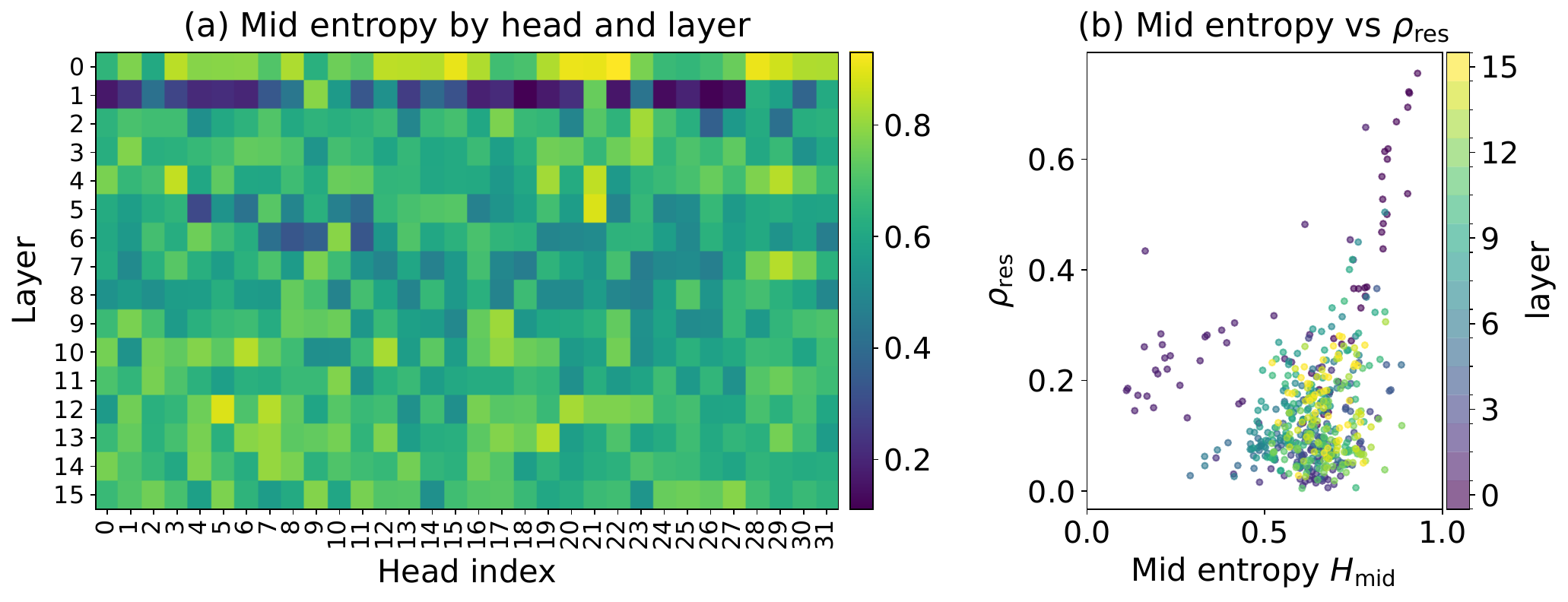}
	\caption{Diffuse-attention and residual-denominator diagnostics for Llama-3.2-1B-Instruct on Wikitext-64k. Panel (a) shows the head/layer map of mid-normalized entropy $H_{\mathrm{mid}}$. Panel (b) plots $H_{\mathrm{mid}}$ against the estimated residual denominator fraction $\rres=\hat Z_{\Rset}/(Z_{\Eset}+\hat Z_{\Rset})$. The fraction $\rres$ depends on the learned $\phi$ residual estimate. It is not the teacher-side mass-recovery quantity $\Cmid(K)$.}
	\label{fig:app_entropy_heatmap}
\end{figure*}

\Cref{fig:app_entropy_heatmap} shows where diffuse mid-region heads occur for Llama-3.2-1B-Instruct on Wikitext-64k. Panel (a) maps mid-normalized entropy over layers and heads. Panel (b) compares entropy with $\rres$. High-entropy heads, roughly $H_{\mathrm{mid}}\gtrsim 0.8$, tend to have larger residual-denominator fractions. This is consistent with the residual-mass interpretation: when mid-region attention is diffuse, a small exact support leaves more mass for the residual branch to account for. The relation is not exact because $\rres$ depends on the learned $\phi$ estimate and is normalized together with anchors and retrieved tokens.

\begin{figure*}[ht]
	\centering
	\includegraphics[width=0.8\linewidth]{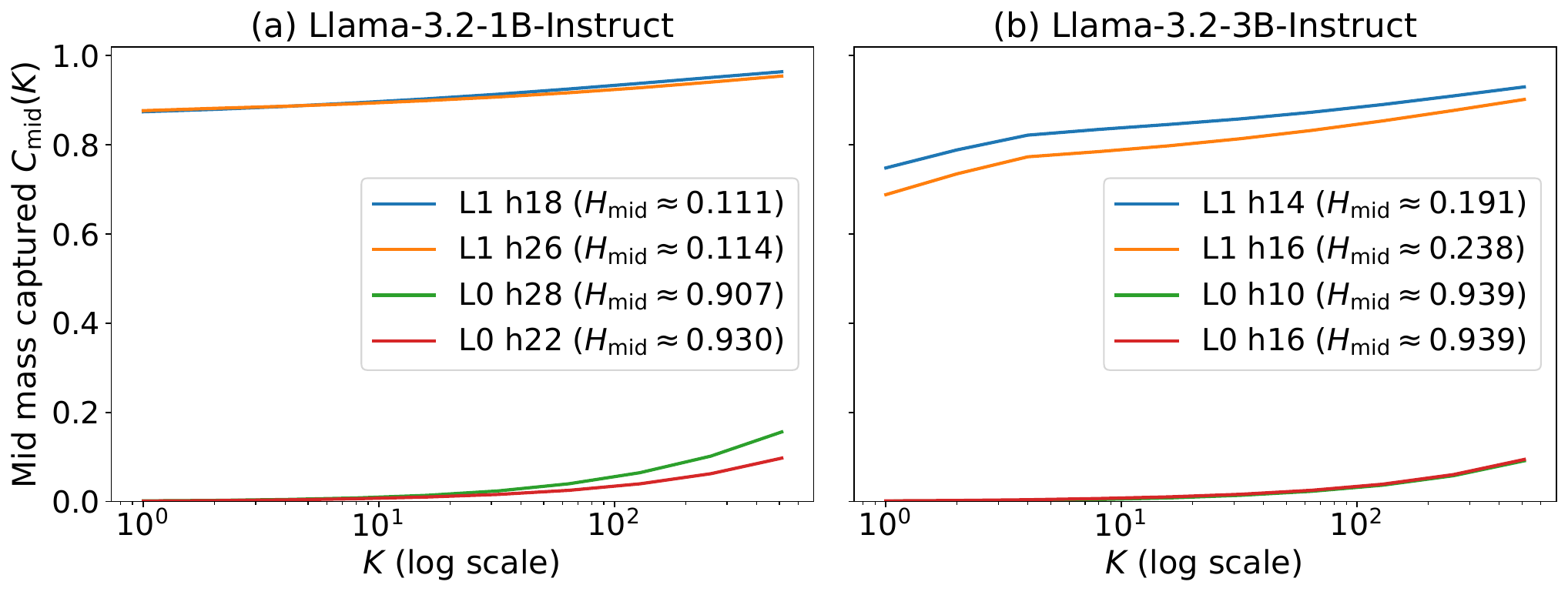}
	\caption{Teacher-side mid-region mass-recovery curves $\Cmid(K)$ for representative heads on Wikitext-64k. Low mid-entropy heads have larger $\Cmid(K)$ at small $K$, meaning that a small number of keys recover most teacher mid-region mass. High mid-entropy heads have smaller $\Cmid(K)$ at the same $K$. Selection-only \TopK{} therefore leaves more teacher mid-region mass unrecovered in those heads.}
	\label{fig:app_mass_curves}
\end{figure*}

\Cref{fig:app_mass_curves} shows teacher-side mass-recovery curves for representative low- and high-entropy heads. Low-entropy heads concentrate mid-region mass on a few keys; consequently, $\Cmid(K)$ rises quickly with small $K$. High-entropy heads spread mass across many keys; consequently, the same $K$ recovers less mid-region mass. This is the regime where selection-only \TopK{} is most likely to discard substantial aggregate mass.

\begin{figure*}[ht]
	\centering
	\includegraphics[width=\linewidth]{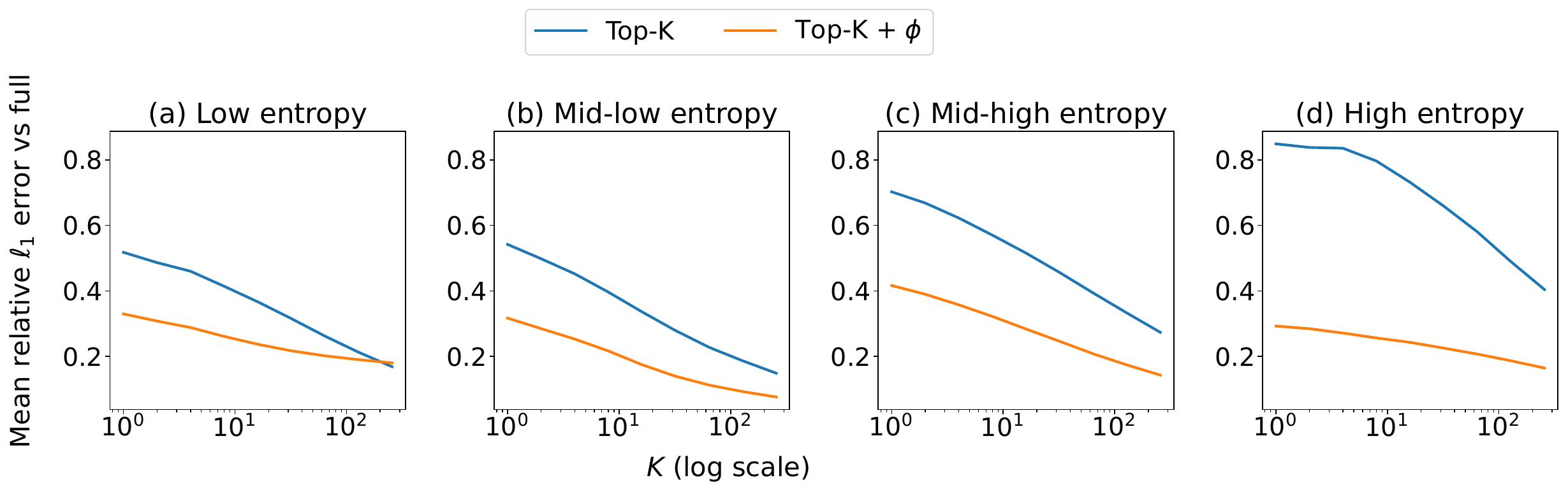}
	\caption{Same-$K$ attention-output error by mid-entropy quartile for Llama-3.2-1B-Instruct on Wikitext-64k. Panels (a)--(d) are ordered from low to high $H_{\mathrm{mid}}$. Each panel compares selection-only \TopK{} and subtractive \TopKphi{} completion against full attention using relative $\ell_1$ error, with the same retrieved set at each $K$.}
	\label{fig:app_quartile_error}
\end{figure*}

\Cref{fig:app_quartile_error} connects missed mid-region mass to attention-output error. Heads are split into mid-entropy quartiles, ordered from low to high entropy. Within each panel, \TopK{} and \TopKphi{} use the same retrieved set at the same $K$; therefore, the comparison isolates residual completion rather than selection. Error generally increases with entropy because the retrieved set captures less of the mid-region mass. \TopKphi{} remains below selection-only \TopK{} across quartiles in this diagnostic, with the largest improvements in the highest-entropy quartiles. As $K$ increases, both methods improve.

\begin{figure*}[ht]
	\centering
	\includegraphics[width=\linewidth]{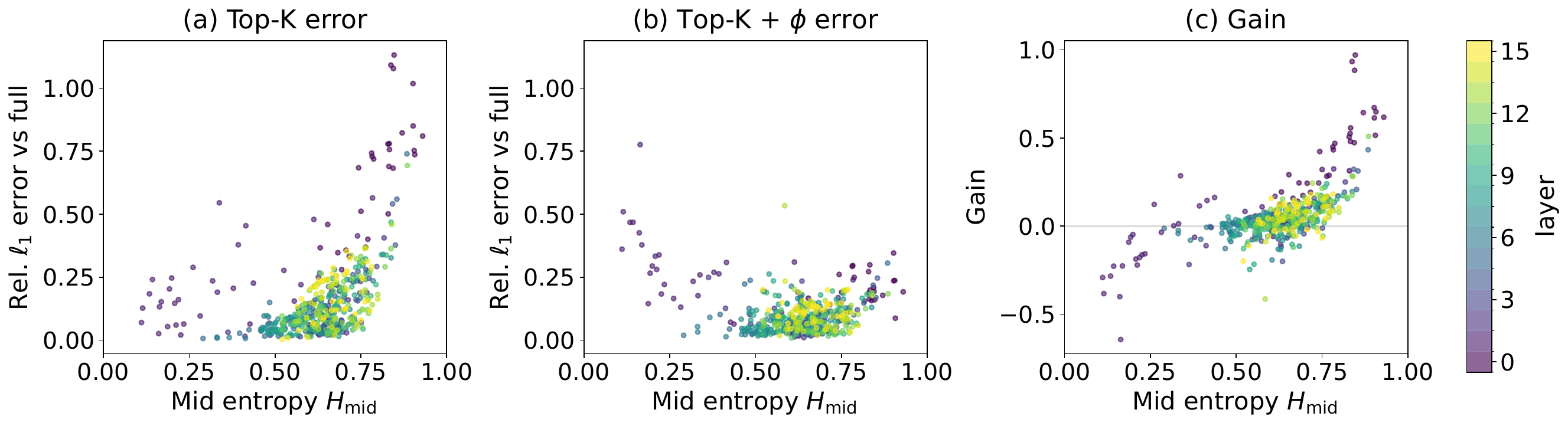}
	\caption{Fixed-$K$ breakdown for Llama-3.2-1B-Instruct on Wikitext-64k. The same retrieved set is used for \TopK{} and \TopKphi{}; therefore, differences isolate the learned residual completion mechanism. Panel (c) plots completion gain $e_{\TopK}-e_{\TopKphi}$ versus $H_{\mathrm{mid}}$, corresponding to the main-text gain diagnostic in \cref{fig:entropy_gain}.}
	\label{fig:app_fixedk_breakdown_1b}
\end{figure*}

\begin{figure*}[ht]
	\centering
	\includegraphics[width=\linewidth]{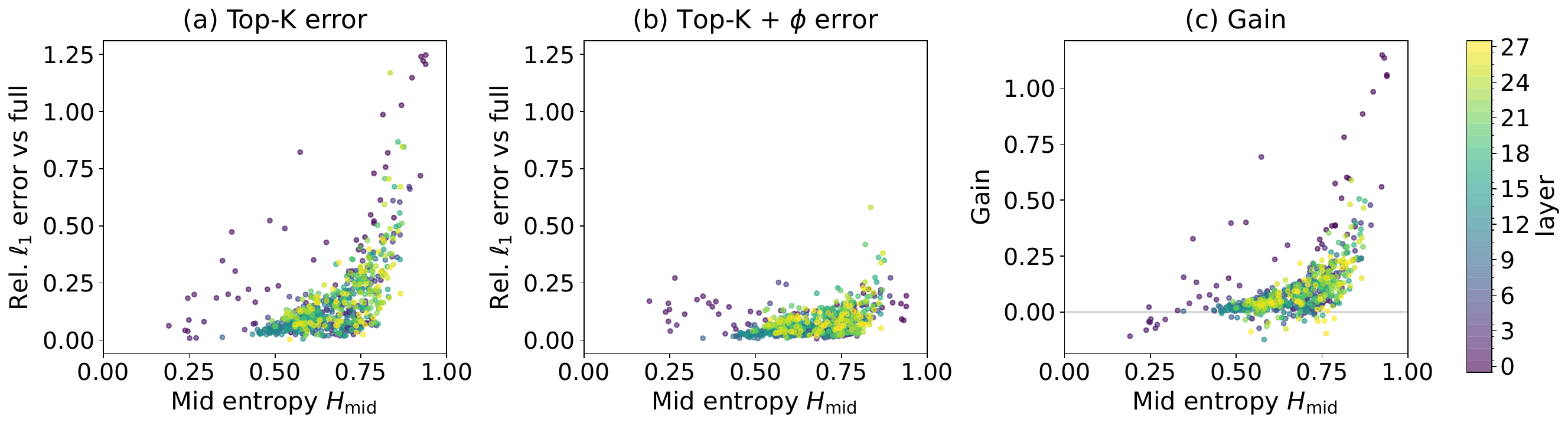}
	\caption{Fixed-$K$ breakdown for Llama-3.2-3B-Instruct on Wikitext-64k, in the same format as \cref{fig:app_fixedk_breakdown_1b}.}
	\label{fig:app_fixedk_breakdown_3b}
\end{figure*}

\Cref{fig:app_fixedk_breakdown_1b,fig:app_fixedk_breakdown_3b} give fixed-$K$ breakdowns for the two Llama backbone LMs. The same retrieved tokens are used by \TopK{} and \TopKphi{}; therefore, the gap again measures the learned residual completion mechanism. Panel (c) in each figure is the fixed-$K$ version of the main completion-gain plot in \cref{fig:entropy_gain}: gain is $e_{\TopK}-e_{\TopKphi}$; hence, positive values indicate that residual completion reduces error.

The figures illustrate a tradeoff. In low-entropy rows, \TopK{} often already captures the dominant mid-region mass; the residual branch has little missing mass to recover, and finite-dimensional $\phi$ error can occasionally make the hybrid output worse. In high-entropy rows, selection-only \TopK{} leaves more diffuse mass unrecovered. The hybrid error also grows in this harder regime, but it grows less than \TopK{} error, producing larger positive gains. Thus, the benefit of completion is strongest when missed mass is diffuse, while the main failure mode is residual-estimation error in rows already well covered by \TopK{}.

\FloatBarrier

\section{Implementation details and third-party assets}
\label{sec:app_reproducibility}

\paragraph{Backbones and evaluation harness.}
We evaluate the frozen Hugging Face Transformers checkpoints \texttt{meta-llama/Llama-3.2-1B-Instruct} and \texttt{meta-llama/Llama-3.2-3B-Instruct}. The evaluation code is based on EleutherAI \texttt{lm-evaluation-harness} v0.4.11. Natural-generation experiments use the LongBench versions of GovReport, QMSum, MuSiQue, and HotpotQA described in the main text, and the controlled long-context evaluations use RULER and BABILong.

\paragraph{Inference settings.}
We use zero-shot evaluation with greedy decoding. The evaluation batch size is 10 for 4k--16k contexts and 1 for 64k contexts and LongBench. All methods use the same 64k input truncation rule described in Section~\ref{sec:experiments}. The exact-support budgets, sink/tail anchor settings, and the definition of $K$ follow Section~\ref{sec:experiments}. Aside from the partial-KV attention replacement and the 64k truncation cap, prompts, generation lengths, and metric computation follow the corresponding \texttt{lm-evaluation-harness} task definitions. We evaluate all harness examples in the reported tasks. RULER uses 500 examples for each of 13 tasks, BABILong uses QA1--QA5 for a total of 4,996 examples, and each LongBench task uses 200 examples.

\paragraph{Hardware and software.}
All main evaluation measurements were run on a single NVIDIA H200 (140.4\,GiB VRAM) attached via PCIe to a host with Intel Xeon Platinum 8488C (2 sockets, 192 CPUs) and 2.0\,TiB system memory. The software stack is Ubuntu 22.04.5 LTS, NVIDIA driver 580.95.05, CUDA 12.8 runtime (driver reports CUDA 13.0), PyTorch 2.9.1+cu128, Transformers 4.57.6, and \texttt{lm-evaluation-harness} v0.4.11. Offline $\phi$ training details and cost are reported in Appendix~\ref{sec:app_phi_training}.

\paragraph{Third-party assets.}
We use publicly available pretrained models, evaluation suites, and corpora, all of which are credited in the main text and bibliography. Third-party assets are used subject to their respective licenses and terms of use.

\FloatBarrier

\clearpage

\end{document}